\pdfoutput=1

\documentclass[11pt, table]{article}

\usepackage{acl}

\usepackage{times}
\usepackage{latexsym}

\usepackage[T1]{fontenc}

\usepackage[utf8]{inputenc}

\usepackage{microtype}

\usepackage{inconsolata}

\usepackage{graphicx}
\usepackage{amsmath, booktabs, multirow,makecell,tabularx}
\usepackage{xcolor} 

\title{DTELS: Towards Dynamic Granularity of Timeline Summarization}

\author{
 \textbf{Chenlong Zhang\textsuperscript{1,2}},
 \textbf{Tong Zhou\textsuperscript{1,2}},
 \textbf{Pengfei Cao\textsuperscript{1,2}},
 \textbf{Zhuoran Jin\textsuperscript{1,2}},
\\
 \textbf{Yubo Chen\textsuperscript{1,2,\thanks{Corresponding Author.}}},
 \textbf{Kang Liu\textsuperscript{1,2,3}},
 \textbf{Jun Zhao \textsuperscript{1,2}}
\\
 \textsuperscript{1}The Key Laboratory of Cognition and Decision Intelligence for Complex Systems,\\Institute of Automation, Chinese Academy of Sciences, Beijing, China,\\
 \textsuperscript{2}School of Artificial Intelligence, University of Chinese Academy of Sciences, Beijing, China,\\
 \textsuperscript{3}Shanghai Artificial Intelligence Laboratory, Shanghai, China
\\
 \small{
   \{zhangchenlong2023, tong$.$zhou\}@ia.ac.cn
 }
 \\
 \small{
 \{pengfei$.$cao, zhuoran.jin, yubo$.$chen, kliu, jzhao\}@nlpr.ia.ac.cn
 }
}

\begin{document}

\maketitle

\begin{abstract}

The rapid proliferation of online news has posed significant challenges in tracking the continuous development of news topics. Traditional timeline summarization constructs a chronological summary of the events but often lacks the flexibility to meet the diverse granularity needs. To overcome this limitation, we introduce a new paradigm, \textbf{D}ynamic-granularity \textbf{T}im\textbf{EL}ine \textbf{S}ummarization, (\textbf{DTELS}), which aims to construct adaptive timelines based on user instructions or requirements. 
This paper establishes a comprehensive benchmark for DTLES that includes: (1) an evaluation framework grounded in journalistic standards to assess the timeline quality across four dimensions: \textit{Informativeness}, \textit{Granular Consistency}, \textit{Factuality}, and \textit{Coherence}; (2) a large-scale, multi-source dataset with multiple granularity timeline annotations based on a consensus process to facilitate authority; (3) extensive experiments and analysis with two proposed solutions based on Large Language Models (LLMs) and existing state-of-the-art TLS methods. The experimental results demonstrate the effectiveness of LLM-based solutions. However, even the most advanced LLMs struggle to consistently generate timelines that are both informative and granularly consistent, highlighting the challenges of the DTELS task.\footnote{Codes are available at \url{https://github.com/chenlong-clock/DTELS-Bench}.}

\end{abstract}

\section{Introduction}
With the surge in news production, the volume of news articles published on the internet is expanding rapidly, making it increasingly challenging to track the developments of news topics. 
\textbf{TimeLine Summarization} (\textbf{TLS}) \citep{wang-etal-2016-low, li-etal-2021-timeline, chen2023follow,zhang2024analyzingtemporalcomplexevents} aims to construct a sequence of chronologically ordered summaries. These timelines provide a traceable skeleton, supporting various applications, including policymaking, crisis management, and stock market analysis.

\begin{figure}[t]
    \centering
    \includegraphics[width=.95\linewidth]{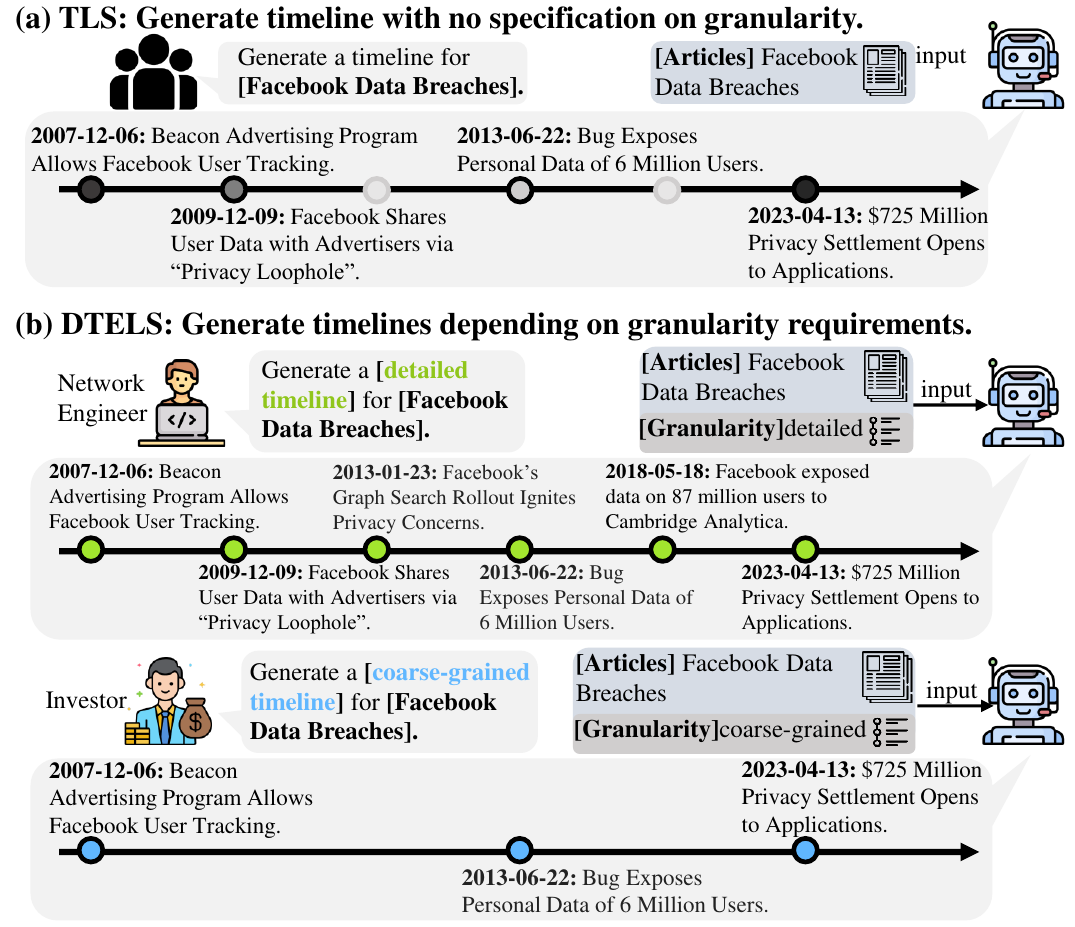}
    \caption{(a) In traditional TLS, a timeline with a predefined number of node summaries is constructed. (b) DTELS provides timelines at different granular levels: network engineers require the technical causes and solutions to data breaches, therefore, a fine-grained granularity is preferred to track the technical details. For investors, a coarse-grained timeline showing the full picture of the breach's influence on investment may suffice.}
    \label{fig:example}
\end{figure}
 
Traditional TLS typically constructs static timelines at a fixed granularity: in Figure \ref{fig:example}a, for a specific news topic, the granularity is heuristically predefined by the number of ``salient events''. However, in practice, the granularity of the timeline should change dynamically, depending on user needs and the nature of news topics:
\textbf{For readers}: Different readers have very different requirements on granularity for the same topic (see example in Figure \ref{fig:example}b). \textbf{For news topics}: A reader's need for granularity varies across topics. One may require fine-grained timelines for trending news such as local disasters to follow the progression and immediate impacts. In contrast, for long-standing topics like the Russian-Ukrainian war, people may warrant coarse-grained timelines with wider intervals to capture broader developments. 

Unfortunately, existing TLS ignores the importance of providing timelines at dynamic granularities.  Existing evaluations also lack appropriate reference annotations and metrics to comprehensively evaluate timelines at dynamic granularities. 

In this paper, we propose a new paradigm: \textbf{D}ynamic granularity \textbf{T}im\textbf{E}\textbf{L}ine \textbf{S}ummarization (\textbf{DTELS}).
We define the granularity of a timeline by the degree of omission between the node summaries.
Given a collection of news articles on the specific news topic and granularity requirements, our task aims to construct dynamic-granularity timelines tailored to various requirements.

Meanwhile, to take the study a step further, grounded in the criteria from journalism \citep{kunelius2006good}, an ideal timeline should:
(1) convey information effectively, avoiding redundant events, and ensuring that no important events are missed.
(2) maintain consistency with the granular requirements.
(3) ensure the mentioned events in each summary are factually correct.
(4) be self-contained, allowing the reader to clearly understand the context.
By adhering to these criteria, we set the standard that not only meets the dynamic granularity needs but also upholds high quality. 

We construct a benchmark including:

\textbf{Evaluation Framework}. To comprehensively measure a timeline, We propose metrics that address the aforementioned criteria: 


\textit{Informativeness}: This metric evaluates the effective volume of information in the node summaries. We propose a ``\textit{mount-then-measure}'' paradigm to align predicted node summaries to those in the reference timeline based on the entailment score of the ``\textit{event atoms}'', which represents the smallest unit of event information within a sentence. 

\textit{Granular Consistency}: The granularity is reflected by the amount of event information omitted between adjacent nodes. The more events omitted, the coarser the granularity is. We regard adjacent nodes as edges and calculate the ratio of mounts on the correct reference granularity edge. 


\textit{Factuality}: Considering the hallucinated contents and misinformation in the era of Large Language Models (LLMs) \citep{ji2023survey,li2023halueval,zhang2023siren}, it is crucial to ensure the information accuracy. We introduce a factuality metric that incorporates atoms-level entailment verification from reference news articles to measure the non-fabricated information in each summary.



\textit{Coherence}: Coherence is pivotal in summarization tasks \citep{goyal-etal-2022-snac, steen-markert-2022-find}. We adopt this metric for our task, ensuring that summaries are generated in a structurally, linguistically, and stylistically coherent manner. To facilitate this, we design a review form to guide the most advanced LLMs for coherence evaluation.

 We verify the effectiveness of the metrics, showcasing high alignment with humans. 

\textbf{Dataset Construction}. To ensure evaluation across varying granularities, we meticulously construct a dataset called \textit{DTELS-Bench}.  We initially collect diverse news topics and journalists' annotations on timelines from news events websites\footnote{\url{https://events.baidu.com} \label{baidu}}. We then gather corresponding large-scale news articles from diverse sources, resulting in a large-scale, multi-source Chinese dataset.
Subsequently, the reference timelines are annotated at three predefined granularities through a consensus-based automated annotation. Finally, the timelines are refined by specialists to ensure the authority.


\textbf{Comprehensive Evaluation}. In the experiments, we present two LLM-based solutions for long-context and context-limited LLMs. We systematically evaluate our proposed solutions with multiple LLMs. In addition, we compare existing state-of-the-art extractive TLS approaches. Experiments show that our LLM-based solutions dominate in all dimensions, however, they fall short of providing high-quality information and aligning the required granularity.
We then analyze the performance of these methods across various settings of DTELS. The results indicate that there is still substantial room for improvement in DTELS. 

To sum up, our contributions are as follows: 
\begin{itemize}
    \item We propose a new task: \textbf{D}ynamic granularity \textbf{T}im\textbf{E}\textbf{L}ine \textbf{S}ummarization (\textbf{DTELS}). It aims to summarize timelines tailored to the unique needs of dynamic granularities. 
    \item We build an event-centric evaluation framework. Extending from journalism, we propose metrics to evaluate timelines in four dimensions: informativeness, granular consistency, factuality, and coherence. Experiments with human annotators demonstrate the effectiveness of our metrics. 
    \item We collect a large-scale, multi-source Chinese dataset, \textit{DTELS-Bench}\footnote{All Chinese information in the paper is translated into English for ease of understanding.}, which contains 543 news topics with 55,432 articles from 2,858 sources. It covers three predefined granularities annotated via a consensus-based mechanism. The expert's refinement enhances the annotation authority.
    \item We evaluate existing state-of-the-art TLS methods as well as LLMs with two proposed DTELS methods. Through extensive experiments, we find the proposed solutions outperform existing TLS methods, however, they are far from being an ideal solution to DTELS.
\end{itemize}

\begin{figure*}[t]
    \centering
    \includegraphics[width=.95\textwidth]{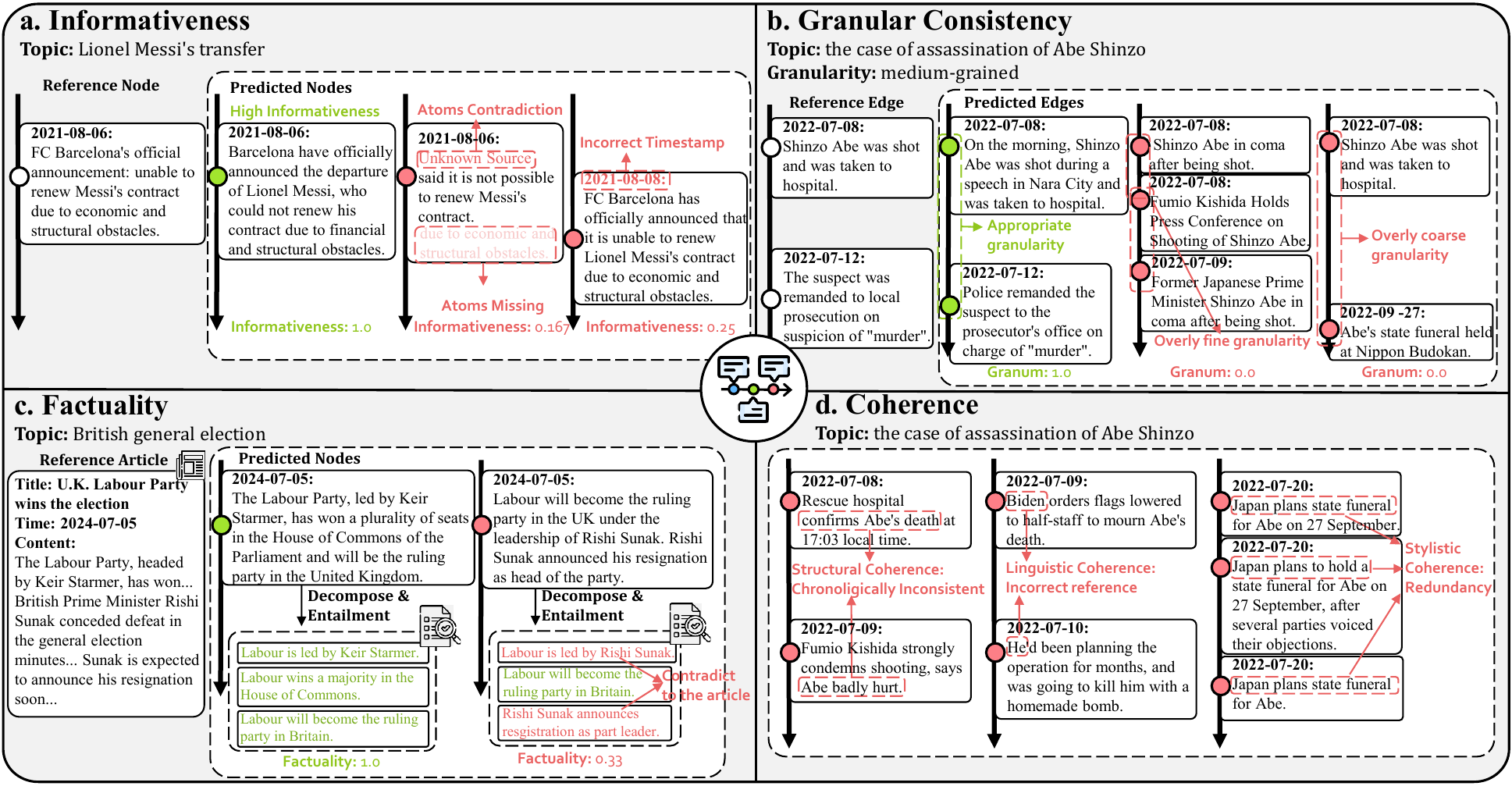}
    \caption{Examples of metrics. Green nodes indicate positive examples and red nodes indicate negative examples.}
    \label{fig:eval_example}
\end{figure*}
\section{Related Works}
\subsection{Timeline Summarization Task}
Timeline Summarization (TLS) has been a long-standing task in Natural Language Processing. 
The challenge of this task is to chronologically condense information from hundreds of articles.
Existing work mainly focuses on generating and evaluating timelines at fixed granularity with sole evaluation metrics. The task is first proposed by \citet{swan2000automatic}. \citet{tran2013leveraging} provides a clear definition of TLS with fixed numbers of nodes for each news topic. \citet{martschat-markert-2017-improving} proposes an alignment-based ROUGE score. However, these works disregard the varying granularity requirements. Besides, the ROUGE-based evaluation \citep{gholipour-ghalandari-ifrim-2020-examining} results can be significantly affected by the narrative styles. 
\subsection{Timeline Summarization Dataset}
\citet{tran2013leveraging} proposes the ``T17'' dataset discussing famous topics. \citet{tran2013leveraging} constructs the ``Crisis'' dataset focusing on long-span armed conflict topics. \citet{wang-etal-2015-socially}. \citet{gholipour-ghalandari-ifrim-2020-examining} builds ``Entities''  with longer time-ranges topics typed around ‘people’ and ‘disasters’. \citet{rajaby-faghihi-etal-2022-crisisltlsum} constructs a dataset called ``CrisisTLS'' focusing on the local crisis. \citet{li-etal-2021-timeline} build a larger dataset TLS\textsubscript{100} covering various topics.  Some recent works also propose LLM-based methods \citep{song-etal-2024-combining,hu-etal-2024-moments,chen-etal-2024-timeline}.
However, they lack annotations across multiple levels. Besides, topics on existing datasets are likely to have been leaked in the pretraining corpus of LLMs, leading to potential unfair evaluations.

\section{Task Definition}
\subsection{Timeline Summarization}
Consider a news topic $\mathbf{q}$ spans over a time range $\mathcal{T} = \{t_1, \dots, t_n\}$ and a corresponding set of news articles $\mathcal{A} = \{A_{t_1}, A_{t_2}, \dots, A_{t_n}\}$ as inputs. Each date $t_i \in \mathcal{T}$ is accompanied by multiple articles $A_{t_i} =\{a_{t_i, 1}, \dots, a_{t_i, m}\}$. The task is to generate a temporal sequence of summaries by model $\Theta$:

\begin{footnotesize}
\begin{equation}
\mathcal{S} = \Theta(\mathbf{q}, \mathcal{A}),
\end{equation}    
\end{footnotesize}

where $\mathcal{S}=\{S_{t_1}, \dots, S_{t_k}\}$ and $k$ corresponds to the node numbers. $S_{t_i}$ includes a timestamp $t_i \in T$ and summary $s_{i}$, i.e., $s_{i}$ is a concise summary of the news event at time $t_i$. Typically, for a specific news topic, the amount of node summaries $k$ is fixed based on the number of salient events.
\subsection{Dynamic-granularity Timeline Summarization} \label{sec:DTELS}
Considering that event information passed through nodes is certain, we define granularity is the degree to which neighboring nodes are omitted: coarse-grained timelines with fewer nodes should omit less important events, while fine-grained timelines capture detailed chronological chains.

We introduce a granularity indicator ``Granularity: [$\mathcal{G}_o$]'' as an additional input to indicate the desired granularity. It can be either a specific number of nodes or a natural language instruction. Here, $m$ denotes the chosen granularity level of the timeline. Based on this, the model $\Theta$ generates a timeline summarization at the specific granularity $\mathcal{G}_o$:

\begin{footnotesize}
    \begin{equation}
    \mathcal{S}^{\mathcal{G}_o} = \Theta(\mathcal{G}_o,\mathbf{q}, \mathcal{A}),
\end{equation}
\end{footnotesize}

where $\mathcal{S}^{\mathcal{G}_o} = \{(t^{\mathcal{G}_o}_1, s^{\mathcal{G}_o}_1), \dots, (t^{\mathcal{G}_o}_k, s^{\mathcal{G}_o}_k)\}$. the granularity of a timeline for a topic can vary: $\mathcal{G} = \{\mathcal{G}_1, \mathcal{G}_2, \dots, \mathcal{G}_n\}$, where $n$ ranges from coarse to fine granularity. This approach ensures that the summarization output matches the specified granularity requirements.
The reference timelines are annotated at multiple granularity levels\footnote{In the following section if $\mathcal{G}_o$ is not explicitly stated, we use the reference timeline with the same granularity $\mathcal{G}_o$ as the predicted timeline for evaluation.}.

\section{Evaluation Framework}
\subsection{Event Atoms} \label{sec:atom}
The references in narrative summarization are influenced by the annotator's preference and narrative style. Existing ROUGE-based evaluation \citep{lin2004ROUGE} approaches evaluate the n-gram similarity, which can be inadequate in fairly reflecting the quality \citep{ng-abrecht-2015-better} of the outputs.

\begin{figure}[t]
    \centering
\includegraphics[width=.9\linewidth]{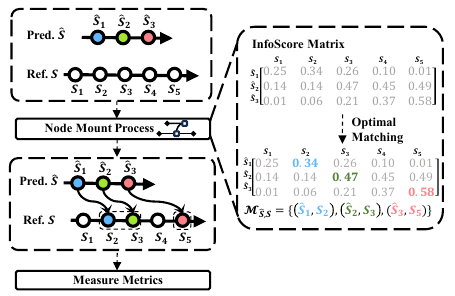}
    \caption{The predicted timeline is mounted to the reference according to ``Optimal Matching''. The colored nodes denote mounted nodes.}
    \label{fig:mount}
\end{figure}

Inspired by recent advances in atom-based evaluations \citep{min-etal-2023-factscore,setty2024factcheck,xu2024faithful}, we introduce the concept of ``event atoms'' as the fundamental units for evaluation,  which remain consistent despite changes in narrative style and granularity. We define the ``event atoms'' as the smallest distinguishable unit of events within a sentence. Each node summary $s_i$ can be decomposed into a certain number of atoms:
$\mathcal{E}_i = \{e_{i,1}, \dots, e_{i,m}\} = Decompose(s_i)$, where $m$ indicates the number of atoms. This function can be achieved by LLMs (detailed in Appendix \ref{app:decompose}).

To evaluate a predicted node summary, we measure the amount of valuable event information it provides compared to the reference node summary using an entailment score. For a predicted node summary $\hat{s}_i$ and a reference node summary $s_j$, their event atoms are $\hat{\mathcal{E}_i}$ and $\mathcal{E}_j$, respectively. The entailment precision $ent_{p}$ can be measured by:

\begin{footnotesize}
\begin{align}
    ent_{p}(\hat{s}_i, s_j) = \frac{1}{|\hat{\mathcal{E}_i}|} \sum_{\hat{\varepsilon}_{i,s} \in \hat{\mathcal{E}_i}}Entail(\mathcal{E}_j, \hat{\varepsilon}_{i,s}),
    \label{ent_p}
\end{align}    
\end{footnotesize}

where event atoms $\hat{\varepsilon}_{i,s}$ derives from $\hat{\mathcal{E}_i}$. $Entail(Evidence, Claim)$ quantifies the entailment of event atoms: it returns 1 if the evidence entails the claim, and 0 if it contradicts or is unrelated to the claim. The function can be implemented by the widely used Natural Language Inference models. \citep{camburu2018snli,klemen-etal-2024-si}. 

Similarly, we can get the entailment recall $ent_{r}(\hat{s}_i, s_j)$. The entailment F1 can be calculated:

\begin{footnotesize}
\begin{align} \label{eq:align_f1}
    ent_{f1}(\hat{s}_i, s_j) = \frac{2 * ent_{p}(\hat{s}_i, s_j)*ent_{r}(\hat{s}_i, s_j) }{ent_{p}(\hat{s}_i, s_j) + ent_{r}(\hat{s}_i, s_j)}.
\end{align}
\end{footnotesize}

By adopting the score, we can evaluate the coverage of the node summaries over the references.

We propose a ``mount-then-measure'' paradigm, illustrated in Figure \ref{fig:mount}, to find the optimal mapping from the predicted timeline to the reference timeline. For a predicted node $\hat{\mathcal{S}}_i = (\hat{t}_i, \hat{s}_i)$, we mount it to a specific reference node $\mathcal{S}_j= (t_j, s_j)$ by computing the information score $InfoScore(\hat{\mathcal{S}}_i, \mathcal{S}_j)$.  Considering the matching of event information on the temporal dimension for timelines, we introduce a temporal interval penalty term $\delta$:

\begin{footnotesize}
\begin{align}
    \delta_{\hat{t}_i, t_j} = \frac{1}{|\hat{t}_i - t_j|^2 + 1}.
\end{align}
\end{footnotesize}
Then, we can define the information score:

\begin{footnotesize}
\begin{align}
    InfoScore(\hat{\mathcal{S}}_i, \mathcal{S}_j) = \delta_{\hat{t_i}, t_j} * ent_{f1}(\hat{s}_i, s_j).
\end{align}
\end{footnotesize}

\subsection{Mount-then-measure Paradigm}
The \textit{InfoScore()} provides an objective measurement of the predicted nodes' coverage from an event-centric perspective. We can get the mapping cost between predicted and reference nodes via:
\begin{align}
    map(\hat{\mathcal{S}}_i \rightarrow \mathcal{S}_j) = -InfoScore(\hat{\mathcal{S}}_i, \mathcal{S}_j).
\end{align}

The mount process for the entire timeline can be automatically completed by Hungarian algorithm \citep{Kuhn1955Hungarian} for a global optimal matching:

\begin{footnotesize}
\begin{align} 
    \mathcal{M}_{\hat{\mathcal{S}}, S} = \arg\min_{\mathcal{M}} \sum_{(\hat{\mathcal{S}}_i, \mathcal{S}_j) \in \mathcal{M}} map(\hat{\mathcal{S}}_i \rightarrow \mathcal{S}_j).
    \label{eq:mount}
\end{align}
\end{footnotesize}
This process determines a maximum coverage of the predicted timeline to the reference, enabling fine-grained evaluation that requires references.


\subsection{Evaluation Metrics}
To evaluate timelines from multiple perspectives, we adopt criteria in journalism \citep{kunelius2006good} and categorize the quality of a timeline into four dimensions: \textit{Informativeness}, \textit{Granular Consistency}, \textit{Factuality}, and \textit{Coherence}. 
The subsequent section details the definition of these metrics.

\textbf{Informativeness.}
Informativeness measures the extent to which the node summary captures the essential information of events. As illustrated in Figure \ref{fig:eval_example}a, it is important to ensure the timeline contains all key atoms at correct timestamps and is not overly verbose. We match references for each node summary by ``mount-then-measure''. We calculate the informativeness $Info()$ after mounting the predicted timeline $\hat{\mathcal{S}}$ to the reference timeline $\mathcal{S}$:

\begin{footnotesize}
\begin{align}
    Info(\hat{\mathcal{S}}) = \frac{1}{|\hat{\mathcal{S}}|} \sum_{(\hat{\mathcal{S}}_i, \mathcal{S}_j) \in \mathcal{M}} InfoScore(\hat{\mathcal{S}}_i, \mathcal{S}_j).
\end{align}    
\end{footnotesize}
$\hat{S}_i$ and $S_j$ are predicted and reference nodes in Equation \ref{eq:mount}. 

\textbf{Granular Consistency.}

Granular consistency measures how well the timeline aligns with its reference in terms of granularity. As illustrated in Figure \ref{fig:eval_example}b, differences in granularity emerge not from individual node content but from relationships between adjacent nodes.

we extend ``mount-then-measure'' to edge views: For a predicted timeline at $\mathcal{G}_o$, its edges are $\hat{\mathbf{E}}=\{\hat{e}_1, \hat{e}_2, \dots, \hat{e}_{k-1}\}$, where $\hat{e}_{m} = (\hat{\mathcal{S}}_m, \hat{\mathcal{S}}_{m+1})$. The reference edges across all granularities are $\mathbf{E}^{\mathcal{G}} = \{\mathbf{E}^{\mathcal{G}_1}, \mathbf{E}^{\mathcal{G}_2}, \dots, \mathbf{E}^{\mathcal{G}_n}\}$.
We calculate the mapping cost of aligning a predicted edge $\hat{e}_m$ to a reference edge $e_n = (\mathcal{S}n, \mathcal{S}{n+1}) \in \mathbf{E}^{\mathcal{G}}$ using the formula:

\begin{footnotesize}
    \begin{equation}
    \begin{aligned} 
    map(\hat{e}_{m} \rightarrow e_{n}) &=-InfoScore(\hat{\mathcal{S}}_m, \mathcal{S}_{n}) \\
    &-InfoScore(\hat{\mathcal{S}}_{m+1}, \mathcal{S}_{n+1}).
    \end{aligned}
\end{equation}
\end{footnotesize}

We then mount $\hat{e}_m$ to a minimum cost $e_n$: 

\begin{footnotesize}
\begin{align}
    \mathcal{M}_{\hat{\mathbf{E}}, \mathbf{E}} = \arg\min_{\mathcal{M}} \sum_{(\hat{e}_m, e_n) \in \mathcal{M}} map(\hat{e}_m \rightarrow e_n).
\end{align}    
\end{footnotesize}
Finally, granular consistency is measured by the number of edges that are aligned with the correct granularity level $\mathcal{G}_o$:

\begin{footnotesize}
\begin{align}
    Granu_i(\hat{\mathcal{S}}) = \frac{1}{|\mathbf{E}|} \sum_{(\hat{e}_m, e_n) \in \mathcal{M}} [e_n \in \mathbf{E}^{\mathcal{G}_o}],
\end{align}
\end{footnotesize}
where $[ ]$ is a binary function.

\textbf{Factuality.}
Factuality measures the faithfulness of summaries, which is crucial given the potential for hallucinated and fabricated content in LLMs \citep{chen-etal-2023-beyond,gekhman-etal-2023-trueteacher}.  In DTELS, factuality assesses whether the information in the timeline can be traced back to support articles. We use a selection mechanism to choose reference articles as support for each predicted node: For a given timestamp $\hat{t}_i$ in the predicted node $\hat{\mathcal{S}}_i$, we select reference articles $\mathcal{A}_{\hat{t}_i}$ that are closest to the timestamp. The factuality score is then computed using entailment precision:

\begin{footnotesize}
\begin{align}
    Fact(\hat{\mathcal{S}}) = \frac{1}{|\hat{\mathcal{S}}|} \sum_{(\hat{s}_i, \hat{t}_i) \in \hat{\mathcal{S}}} ent_p(\hat{s}_{i}, \mathcal{A}_{\hat{t}_i}).
\end{align}
    
\end{footnotesize}
The articles are decomposed into a set of event atoms $\mathcal{E}_A$ as reference event atoms in equation \ref{ent_p}. If the node summary contains hallucinated or fabricated content, it won't be fully entailed by the reference articles (see Figure \ref{fig:eval_example}c).

\textbf{Coherence.}
\label{sec:coherence}
While coherence is crucial in document summarization tasks \citep{wu2018learning,chang2024booookscore}, directly applying it to timeline summarization is insufficient. Unlike standard summaries that emphasize narrative coherence, timeline summaries demand structural coherence, including linguistic and stylistic consistency.

Figure \ref{fig:eval_example}d shows common coherence issues. We introduce an evaluation process similar to the ACL Review Form\footnote{\url{https://aclrollingreview.org/reviewform}}, assessing Structural, Linguistic, and Style Coherence, with details in Appendix \ref{app:coherence}.

The process involves: (1) Paraphrasing content to improve understanding and reduce bias; (2) Rating each aspect from 1 to 3 and explaining the rationale for fine-grained evaluation; (3) Giving an overall score from 1 to 5 for a holistic assessment.

To reduce reviewers' workload, we use GPT-4o API\footnote{\url{https://platform.openai.com/docs/models/GPT-4o} \label{gpt4}} for automatic coherence assessment. Domain experts provide annotated examples to guide the model in understanding the criteria.

\section{Dataset Construction} \label{sec:data}
To ensure comprehensive evaluation across timelines at different granularity levels, dataset construction must meet two key premises: (1) The dataset should include news topics of varying complexity, types, and scales, with articles from diverse sources to simulate different granularity needs, enabling robust evaluation. (2) During annotations, annotators should minimize personal biases and annotate nodes at multiple granularity levels to facilitate the evaluation of both fine-grained and coarse-grained timelines. Our solutions to these challenges will be discussed in the following sections.

\begin{table}[t]
    \centering
    \setlength{\tabcolsep}{1mm} 
    \small
    \resizebox{0.9\linewidth}{!}{
         \begin{tabular}{l c c c c c c}
    
        \toprule
        \multirow{2}{*}{\textbf{Dataset}} & \multirow{2}{*}{\textbf{\makecell{\#Topics}}} & \multirow{2}{*}{\textbf{\makecell{\#Topic\\types}}} & \multirow{2}{*}{\textbf{\#Articles}} & \multirow{2}{*}{\textbf{\#Sources}} & \multirow{2}{*}{\textbf{\#Granu}} \\
        \\
        \midrule
        T17 & 9 & 1 & 4,650 & 2 & 1 \\
        Crisis & 4 & 1 & 9,240 & 3 & 1 \\
        Entities & 47 & 2 & 45,075 & 1 & 1 \\ 
        CrisisTL & \textbf{1,000} & 1 & 10,610 & 1 & 1 \\
    TLS\textsubscript{100} & 100 & 4 & 10,379 & 2 & 1  \\
        \midrule
        Ours & 543 & \textbf{7} & \textbf{55,432} & \textbf{2,858} & \textbf{3} \\
        \bottomrule
    \end{tabular}
    }
   
    \caption{Comparison with existing datasets.}
    \label{tab:dataset}
\end{table}

\subsection{Data Collection}



For data collection, we aim to assemble a diverse and representative set of news topics. We begin by leveraging Baidu's event news websites, known for expert fine-grained timeline annotation, to obtain news topics and their corresponding reference timelines. We then manually filter these to ensure quality and diversity based on a standard.

The final dataset includes 543 news topics after October 2023, categorized into seven major types (Politics, Economy, Society, Science, Technology, Sports, and Entertainment), with reference timelines ranging from 9 to 200 nodes.

To gather reference articles, we use Baidu, Google, and Bing, employing multiple keywords to ensure each node is supported by at least 5 articles on average. The final dataset includes 55,432 articles, averaging 102 articles per topic. Table \ref{tab:dataset} compares our dataset with existing datasets.



\begin{figure}[t]
    \centering
    \includegraphics[width=0.95\linewidth]{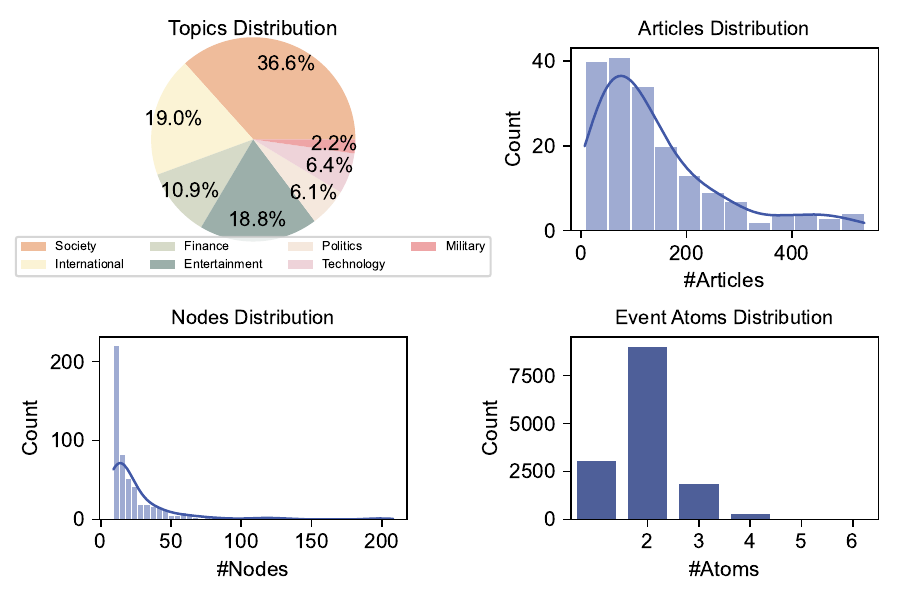}
    \caption{Dataset statistics.}
    \label{fig:dataset}
\end{figure}

\begin{table*}[t]
    \centering
    \small
    \setlength{\tabcolsep}{1mm} 
    \resizebox{.90\textwidth}{!}{
        \begin{tabular}{l c c c c c c c c c c c c c c}
        \toprule
        \multirow{2}{*}{\makecell{Methods}} & \multirow{2}{*}{Models} & \multicolumn{4}{c}{Granularity $\mathcal{G}_N$} & \multicolumn{4}{c}{Granularity $\mathcal{G}_{10}$} & \multicolumn{4}{c}{Granularity $\mathcal{G}_5$} \\
        \cmidrule(lr){3-6} \cmidrule(lr){7-10} \cmidrule(lr){11-14}
        & & Info & Granu\textsubscript{N} & Fact & Coherence & Info & Granu\textsubscript{10} & Fact & Coherence & Info & Granu\textsubscript{5} & Fact & Coherence\\
        \midrule
        Datewise & - & \textbf{17.35} & \textbf{79.15} & \textbf{76.7} & \textbf{56.27} & \textbf{5.2} & 16.13 & \textbf{74.37} & \textbf{55.21} & \textbf{4.46} & 8.06 & \textbf{72.24} & \textbf{57.99} \\
        Cluster & - & 4.14 & 72.01 & 69.6 & 55.33 & 2.65 & \textbf{16.90} & 66.27 & 52.56 & 2.32 & \textbf{10.73} & 64.79 & 54.10 \\
        \midrule
        TO & GPT-3.5-Turbo & 1.45 & 60.21 & 41.19 & \textbf{\underline{91.20}} & 0.83 & 20.54 & 41.22 & \textbf{\underline{94.76}} & 0.65 & 11.58 & 38.99 & \textbf{\underline{97.46}} \\
        \midrule
        \multirow{3}{*}{LP} & GPT-4o & 6.55 & 61.94 & \textbf{65.78} & \textbf{69.21} & 0.92 & 8.80 & \textbf{86.82} & \textbf{77.15} & 0.74 & 3.99 & \textbf{88.11} & \textbf{87.55} \\
        & GLM-3-Turbo & 1.51 & 56.16 & 45.04 & 60.20 & 4.45 & 20.64 & 71.84 & 62.99 & 4.69 & 11.51 & 70.58 & 70.95 \\
        & Yi-medium & \textbf{9.87} & \textbf{66.45} & 65.39 & 63.10 & \textbf{\underline{4.91}} & 17.91 & 77.48 & 65.49 & \textbf{\underline{8.69}} & \textbf{\underline{23.36}} & 51.32 & 71.88 \\
        \rowcolor{gray!25} 

        LP\textsubscript{GT}& GPT-4o & 1.91 & 59.69 & 48.24 & 55.89 & 2.17 & \textbf{\underline{26.74}} & 46.56 & 56.28 & 1.76 & 14.78 & 47.94 & 56.30 \\
        \midrule
        \midrule
        \multirow{7}{*}{HM} & GPT-3.5-Turbo & 24.24 & \textbf{\underline{81.72}} & 91.95 & 65.87 & 0.82 & 7.96 & 91.96 & 68.38 & 0.72 & 4.74 & 91.96 & 76.92 \\
        & GLM-3-Turbo & 21.07 & 72.43 & 87.39 & 67.40 & 1.37 & \textbf{15.65} & 87.61 & 68.01 & 0.91 & \textbf{8.74} & 88.33 & 71.34 \\
        & Yi-medium & 17.46 & 75.32 & 82.26 & 64.28 & \textbf{2.36} & 14.34 & 86.02 & 65.56 & 1.75 & 6.91 & 85.60 & 73.41 \\
        & Qwen1.5-110b & 28.00 & 76.51 & 83.99 & \textbf{78.36} & 2.24 & 10.75 & 83.27 & 79.77 & \textbf{1.78} & 6.81 & 81.09 & \textbf{86.69} \\
        & Qwen1.5-72b &  24.69 & 80.25 &  85.14 &       74.82 &   0.92 &         10.31 &  85.4  &       \textbf{80.86} &   0.74 &          5.48 &  84.56 &       85.57 \\
        & Qwen1.5-32b &  23.37 & 73.97 &  86.29 &       68.64 &   0.61 &         10.14 &  86.32 &       75.47 &   0.55 &          5.54    &  88.04 &       82.08 \\
        & Qwen1.5-14b &  25.26 & 67.78 &  85.76 &       69.69 &   0.71 &         13.06 &  86.31 &       69.98 &   0.56 &          6.98 &  85.58 &       78.64  \\
        \rowcolor{gray!25} 
        HM\textsubscript{GT}& GPT-3.5-Turbo & \textbf{\underline{36.82}} & 78.59 & \textbf{\underline{94.63}} & 64.20 & 1.21 & 9.41 & \textbf{\underline{93.59}} & 70.00 & 1.02 & 6.07 & \textbf{\underline{93.48}} & 68.60 \\
        \bottomrule
    \end{tabular}
    }
    \caption{Main results of different methods on DTELS task. The best results for different methods are in \textbf{bold}. The best results across all methods are \underline{underlined}.}
    \label{tab:main_results}
\end{table*}

\subsection{Consesus-based Annotation}

DTELS requires multiple granularity levels, but annotating all is impractical. To aid evaluation, we define three levels: fine-grained (${G_N}$), medium-grained (${G_{10}}$), and coarse-grained (${G_5}$), where N, 10, and 5 denote the number of nodes in the reference timelinee\footnote{For simplicity, granularity is defined by node quantity. In application, the node numbers do not affect evaluation.}. ${G_N}$ reflects the original timeline with an unspecified node count, while medium and coarse timelines are annotated through consensus.


Even experienced journalists may differ in selecting events for coarse-grained timelines from fine-grained ones. To ensure uniformity, maximizing consensus among annotators on salient events and granularity is essential. However, this process can be costly and time-consuming in DTELS, especially with numerous articles per topic. We utilize GPT-4o to facilitate consensus through role-playing \citep{he-etal-2023-lego,tao-etal-2024-chatgpt}.
The consensus-based annotation process involves three stages:
(1) \textit{Salient Events Decomposition}: For medium-grained timelines $\mathcal{S}^{\mathcal{G}_{10}}$, we decompose the fine-grained timeline $\mathcal{S}^{\mathcal{G}_N}$ into event atoms and group them by timestamp.
(2) \textit{Consensus-based Selection}: For each news topic, we prompt GPT-4o in different roles to select the 10 most important event groups from the atom groups, based on consensus among three roles.
(3) \textit{Expert Refinement}: Domain experts refine the selected groups to ensure quality, summarizing them into a 10-node timeline. The fine-grained timelines are annotated similarly.
We show details and agreement in Appendix \ref{app:consensus}. The results with high inter-annotator agreement show the effectiveness of our annotation.

We list the statistics of the dataset in Figure \ref{fig:dataset}. A more detailed description of dataset construction and annotation can be found in Appendix \ref{app:dataset}.



\section{Experiments} \label{exp}

\subsection{Experimental Settings}
For \textbf{extractive methods}, We implement two state-of-the-art methods \citep{gholipour-ghalandari-ifrim-2020-examining} as baselines: 

\textbf{Datewise}: This method selects key dates in a regression-based manner and then applies centroid-opt \citep{gholipour-ghalandari-2017-revisiting} to extract summaries for each date. 

\textbf{Clustering}: 
This method clusters articles using TF-IDF vectors and then converts the clusters into a temporal graph. Dates are assigned to each cluster through a regression model. For DTELS, we constrain the number of nodes in the timeline according to the specified granularity level.

For \textbf{LLMs}, we select LLMs with Chinese ability. We propose two solutions: 

\textbf{Long-context Prompting} (\textbf{LP}): For the long-context model, we directly prompt the model by providing the news topic, the entire articles with timestamps, and the granularity instruction.

\textbf{Hierarchical Merging} (\textbf{HM}): For models with limited context length, they generate summaries for each date according to the input articles's timestamps. Subsequently, these summaries are hierarchically merged following the merging prompts.

We also establish two distinct experimental settings to evaluate the task's characteristics: 

\textbf{Gold Timestamps} (\textbf{GT}): We instruct models with correct timestamps to guide content generation, ensuring the focus is on content quality rather than timestamp accuracy. This setting can be used for both LP (LP\textsubscript{GT}) and HM (HM\textsubscript{GT}).

\textbf{Topic Only} (\textbf{TO}): Only providing the news topics and granularity requirements to generate a fabricated timeline. 

The full implementations of the methods and model are detailed in Appendix \ref{app:method}.

\subsection{Main Results}


We conduct experiments with the proposed metrics. The main results are shown in Table \ref{tab:main_results}. From the results, we can observe the following conclusions:

\textit{LLMs dominate in DTELS}. 
LLM-based methods outperform state-of-the-art models across all metrics. The HM excels at $\mathcal{G}_N$ in \textit{Info} and \textit{Granu}. while LP performs robustly at coarse and medium granularities, indicating the hierarchical method's strength in capturing details and LP's capability in managing timelines with long-context windows.

\textit{Context window matters for long-context prompting}. With 200k context windows, Yi-medium-200k outperforms models with 128k windows, particularly at coarse granularities, demonstrating the effectiveness in broad event overviews.

\textit{Model capacity influences fine-grained metrics}. Results from Qwen at different scales show that as model size decreases, performance in informativeness and granular consistency declines, suggesting that larger models are better at capturing and conveying detailed information.

\textit{Observation from the variants}. With gold timestamps in ``HM'', the factuality is enhanced with temporal guidance. However, timestamps provide minimal benefits to LP and may reduce factuality and coherence. The ``Topic Only'' approach achieves the highest coherence scores but significantly lags in factuality, indicating that it maintains narrative continuity at the cost of factual accuracy.

\section{Analysis}
To further assess performance across different granularities, we conduct an extended evaluation using hierarchical merging with GPT-3.5-Turbo. 
\subsection{Extended Evaluation on Granularities}
\begin{figure}[t]
    \centering
\includegraphics[width=1.0\linewidth]{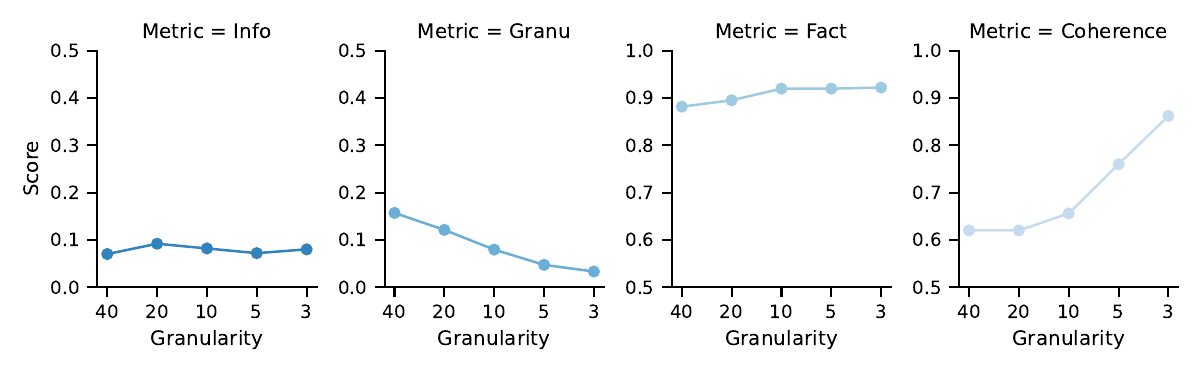}
    \caption{
    Extended evaluation on granularity levels.
    }
    \label{fig:more_granularity}
\end{figure}

\textbf{More Detailed Granularities.}
We define five distinct granularity levels: $\mathcal{G}_{40}$, $\mathcal{G}_{20}$, $\mathcal{G}_{10}$, $\mathcal{G}_{5}$, and $\mathcal{G}_3$. We collect a subset of the dataset with each topic containing over 50 nodes at $\mathcal{G}_N$. Reference timelines are annotated at these levels, and performance is assessed using Hierarchical Merging with GPT-3.5-Turbo. Results, shown in Figure \ref{fig:more_granularity}, reveal that while coarser summaries generally offer better informativeness, factuality, and coherence, they may struggle with granular consistency, highlighting a trade-off between detail and summary quality.

\begin{table}[t]
    \centering
    \small
    \setlength{\tabcolsep}{1mm} 
    \resizebox{.9\linewidth}{!}{
    \begin{tabular}{cccccc}
    \hline
     Granularity  &  \makecell{Granular\\Instruction}  &   Info &   Granu\textsubscript{i} &   Fact &   Coherence \\
    \hline
    \multirow{4}{*}{$\mathcal{G}_N$} & Prompt   &   10.10    &         75.61 &  91.67 & 68.42 \\
    & One-shot &  22.65 &  81.10  &  \textbf{93.88} &        \textbf{70.86} \\
    \cmidrule{2-6}
    & \#Node  &  \textbf{24.24} &         \textbf{81.72}  &  91.95 &        65.87 \\
    \midrule
    \multirow{4}{*}{$\mathcal{G}_5$} & Prompt & 0.74 &          7.26 &  91.44 &        69.2 \\
    & One-shot & \textbf{0.75} &          \textbf{7.41} &  \textbf{92.37} &        71.79 \\
    \cmidrule{2-6}
    & \#Node & 0.72 &          4.74 &  91.96 &        \textbf{76.92} \\
    \hline
    \end{tabular}
    }
    \caption{
    Results of natural language granularity instructions, where \#Node represents the HM method with the number of nodes as the granularity instruction.
    }
    \label{tab:natural_instruct_result}
\end{table}

\textbf{Natural Language Granularity Instructions}
We evaluate natural language granularity instructions, defining fine- and coarse-grained instructions (see Table \ref{tab:natural_instruct_result}). Reference timelines include the both fine- ($\mathcal{S}^{\mathcal{G}_N}$) and coarse-grained ($S^{G_5}$). Results in Table \ref{tab:natural_instruct_result} show that the one-shot method performs competitively with the \#Node method, indicating that models learn to generate accurate timelines with natural language granularity instructions.
\subsection{Influential Factors on Metrics}
We analyze the influence of topics and the article numbers. We conclude that the two aspects greatly influence the performance. Results in Appendix \ref{app:exp} suggest improvement in stability is necessary.

\subsection{Metrics Alignment with Human}

\textbf{Agreement Score.}
Annotators are asked to rate timelines on a scale from 1 to 5 based on the metrics. Pearson correlations are: informativeness (78.74\%), granular consistency (76.66\%), factuality (95.87\%), and coherence (99.14\%).

\textbf{Consistency Score.}
Given pairs of timelines for a topic generated by two models, annotators rate the better one for each metric. We calculate the consistent score between annotators and metrics. Each metric's consistency exceeds 90\%, showing high consistency between humans and the metrics.

Details can be found in the Appendix \ref{app:alignment}.
\section{Conclusions}

In this paper, we introduce a \textbf{D}ynamic-granularity \textbf{T}im\textbf{EL}ine \textbf{S}ummarization (\textbf{DTELS}) task, which aims to construct timeline summaries at dynamic granularity levels following the granularity requirements. We build a comprehensive benchmark including: 
(1) \textbf{Evaluation Framework}: We propose an event-centric evaluation along with metrics: informativeness, granular consistency, factuality, and coherence. Evaluation of alignment with the human annotator proves the rationality of the proposed metrics. (2) \textbf{Dataset Construction}: We construct a large-scale Chinese dataset for DTELS with consensus-based annotation for multi-granularity references. We apply expert refinement to ensure the authority of the annotation. (3) \textbf{Comprehensive Evaluation}: We present two solutions for large language models. Through experiments on existing state-of-the-art timeline summarization methods as well as LLM-based solutions on multiple models, we find that the DTELS task remains challenging. Further research is required to improve the informativeness granularity consistency. In the future, we plan to diversify the language sources and improve LLM-based methods to better capture information and enhance granular consistency.

\section*{Limitations}
Though our DTELS approach has shown promising results, there are several limitations that need to be addressed in future work: Our approach relies heavily on the availability of a large-scale, annotated dataset. The creation of such datasets is time-consuming, which may limit the scalability and applicability of our approach to other domains or languages where such resources are not available. To evaluate the generated timelines, we rely on large language models' APIs, which are costly and may not be accessible to all researchers. Besides, The language of our dataset is Chinese, which may limit the generalizability of our approach to other languages. Further research is needed to develop more efficient data collection and evaluation methods that can be applied to a wider range of languages and domains.

\bibliography{custom}

\begin{thebibliography}{40}
\providecommand{\natexlab}[1]{#1}

\bibitem[{Achiam et~al.(2023)Achiam, Adler, Agarwal, Ahmad, Akkaya, Aleman, Almeida, Altenschmidt, Altman, Anadkat et~al.}]{achiam2023gpt}
Josh Achiam, Steven Adler, Sandhini Agarwal, Lama Ahmad, Ilge Akkaya, Florencia~Leoni Aleman, Diogo Almeida, Janko Altenschmidt, Sam Altman, Shyamal Anadkat, et~al. 2023.
\newblock Gpt-4 technical report.
\newblock \emph{arXiv preprint arXiv:2303.08774}.

\bibitem[{Camburu et~al.(2018)Camburu, Rockt{\"a}schel, Lukasiewicz, and Blunsom}]{camburu2018snli}
Oana-Maria Camburu, Tim Rockt{\"a}schel, Thomas Lukasiewicz, and Phil Blunsom. 2018.
\newblock e-snli: Natural language inference with natural language explanations.
\newblock \emph{Advances in Neural Information Processing Systems}, 31.

\bibitem[{Chang et~al.(2024)Chang, Lo, Goyal, and Iyyer}]{chang2024booookscore}
Yapei Chang, Kyle Lo, Tanya Goyal, and Mohit Iyyer. 2024.
\newblock \href {https://openreview.net/forum?id=7Ttk3RzDeu} {Booookscore: A systematic exploration of book-length summarization in the era of {LLM}s}.
\newblock In \emph{The Twelfth International Conference on Learning Representations}.

\bibitem[{Chen et~al.(2024)Chen, Ouyang, Ren, Ding, Hu, and Qu}]{chen-etal-2024-timeline}
Jianhao Chen, Haoyuan Ouyang, Junyang Ren, Wentao Ding, Wei Hu, and Yuzhong Qu. 2024.
\newblock \href {https://aclanthology.org/2024.acl-long.187} {Timeline-based sentence decomposition with in context learning for temporal fact extraction}.
\newblock In \emph{Proceedings of the 62nd Annual Meeting of the Association for Computational Linguistics (Volume 1: Long Papers)}, pages 3415--3432, Bangkok, Thailand. Association for Computational Linguistics.

\bibitem[{Chen et~al.(2023{\natexlab{a}})Chen, Deng, Bian, Qin, Wu, Chua, and Wong}]{chen-etal-2023-beyond}
Liang Chen, Yang Deng, Yatao Bian, Zeyu Qin, Bingzhe Wu, Tat-Seng Chua, and Kam-Fai Wong. 2023{\natexlab{a}}.
\newblock \href {https://doi.org/10.18653/v1/2023.emnlp-main.390} {Beyond factuality: A comprehensive evaluation of large language models as knowledge generators}.
\newblock In \emph{Proceedings of the 2023 Conference on Empirical Methods in Natural Language Processing}, pages 6325--6341, Singapore. Association for Computational Linguistics.

\bibitem[{Chen et~al.(2023{\natexlab{b}})Chen, Li, Gao, Chan, Zhao, Gao, Zhang, and Yan}]{chen2023follow}
Xiuying Chen, Mingzhe Li, Shen Gao, Zhangming Chan, Dongyan Zhao, Xin Gao, Xiangliang Zhang, and Rui Yan. 2023{\natexlab{b}}.
\newblock Follow the timeline! generating an abstractive and extractive timeline summary in chronological order.
\newblock \emph{ACM Transactions on Information Systems}, 41(1):1--30.

\bibitem[{Devlin et~al.(2018)Devlin, Chang, Lee, and Toutanova}]{devlin2018bert}
Jacob Devlin, Ming-Wei Chang, Kenton Lee, and Kristina Toutanova. 2018.
\newblock Bert: Pre-training of deep bidirectional transformers for language understanding.
\newblock \emph{arXiv preprint arXiv:1810.04805}.

\bibitem[{Gekhman et~al.(2023)Gekhman, Herzig, Aharoni, Elkind, and Szpektor}]{gekhman-etal-2023-trueteacher}
Zorik Gekhman, Jonathan Herzig, Roee Aharoni, Chen Elkind, and Idan Szpektor. 2023.
\newblock \href {https://doi.org/10.18653/v1/2023.emnlp-main.127} {{T}rue{T}eacher: Learning factual consistency evaluation with large language models}.
\newblock In \emph{Proceedings of the 2023 Conference on Empirical Methods in Natural Language Processing}, pages 2053--2070, Singapore. Association for Computational Linguistics.

\bibitem[{Gholipour~Ghalandari(2017)}]{gholipour-ghalandari-2017-revisiting}
Demian Gholipour~Ghalandari. 2017.
\newblock \href {https://doi.org/10.18653/v1/W17-4511} {Revisiting the centroid-based method: A strong baseline for multi-document summarization}.
\newblock In \emph{Proceedings of the Workshop on New Frontiers in Summarization}, pages 85--90, Copenhagen, Denmark. Association for Computational Linguistics.

\bibitem[{Gholipour~Ghalandari and Ifrim(2020)}]{gholipour-ghalandari-ifrim-2020-examining}
Demian Gholipour~Ghalandari and Georgiana Ifrim. 2020.
\newblock \href {https://doi.org/10.18653/v1/2020.acl-main.122} {Examining the state-of-the-art in news timeline summarization}.
\newblock In \emph{Proceedings of the 58th Annual Meeting of the Association for Computational Linguistics}, pages 1322--1334, Online. Association for Computational Linguistics.

\bibitem[{Goyal et~al.(2022)Goyal, Li, and Durrett}]{goyal-etal-2022-snac}
Tanya Goyal, Junyi~Jessy Li, and Greg Durrett. 2022.
\newblock \href {https://doi.org/10.18653/v1/2022.emnlp-main.29} {{SN}a{C}: Coherence error detection for narrative summarization}.
\newblock In \emph{Proceedings of the 2022 Conference on Empirical Methods in Natural Language Processing}, pages 444--463, Abu Dhabi, United Arab Emirates. Association for Computational Linguistics.

\bibitem[{He et~al.(2023)He, Cao, Chen, Liu, Li, Sun, and Zhao}]{he-etal-2023-lego}
Zhitao He, Pengfei Cao, Yubo Chen, Kang Liu, Ruopeng Li, Mengshu Sun, and Jun Zhao. 2023.
\newblock \href {https://doi.org/10.18653/v1/2023.findings-emnlp.613} {{LEGO}: A multi-agent collaborative framework with role-playing and iterative feedback for causality explanation generation}.
\newblock In \emph{Findings of the Association for Computational Linguistics: EMNLP 2023}, pages 9142--9163, Singapore. Association for Computational Linguistics.

\bibitem[{Hu et~al.(2024)Hu, Moon, and Ng}]{hu-etal-2024-moments}
Qisheng Hu, Geonsik Moon, and Hwee~Tou Ng. 2024.
\newblock \href {https://aclanthology.org/2024.acl-long.390} {From moments to milestones: Incremental timeline summarization leveraging large language models}.
\newblock In \emph{Proceedings of the 62nd Annual Meeting of the Association for Computational Linguistics (Volume 1: Long Papers)}, pages 7232--7246, Bangkok, Thailand. Association for Computational Linguistics.

\bibitem[{Ji et~al.(2023)Ji, Lee, Frieske, Yu, Su, Xu, Ishii, Bang, Madotto, and Fung}]{ji2023survey}
Ziwei Ji, Nayeon Lee, Rita Frieske, Tiezheng Yu, Dan Su, Yan Xu, Etsuko Ishii, Ye~Jin Bang, Andrea Madotto, and Pascale Fung. 2023.
\newblock Survey of hallucination in natural language generation.
\newblock \emph{ACM Computing Surveys}, 55(12):1--38.

\bibitem[{Klemen et~al.(2024)Klemen, {\v{Z}}agar, {\v{C}}ibej, and Robnik-{\v{S}}ikonja}]{klemen-etal-2024-si}
Matej Klemen, Ale{\v{s}} {\v{Z}}agar, Jaka {\v{C}}ibej, and Marko Robnik-{\v{S}}ikonja. 2024.
\newblock \href {https://aclanthology.org/2024.lrec-main.1294} {{SI}-{NLI}: A {S}lovene natural language inference dataset and its evaluation}.
\newblock In \emph{Proceedings of the 2024 Joint International Conference on Computational Linguistics, Language Resources and Evaluation (LREC-COLING 2024)}, pages 14859--14870, Torino, Italia. ELRA and ICCL.

\bibitem[{Kuhn(1955)}]{Kuhn1955Hungarian}
Harold~W. Kuhn. 1955.
\newblock \href {https://doi.org/10.1002/nav.3800020109} {{The Hungarian Method for the Assignment Problem}}.
\newblock \emph{Naval Research Logistics Quarterly}, 2(1--2):83--97.

\bibitem[{Kunelius(2006)}]{kunelius2006good}
Risto Kunelius. 2006.
\newblock Good journalism: On the evaluation criteria of some interested and experienced actors.
\newblock \emph{Journalism studies}, 7(5):671--690.

\bibitem[{Li et~al.(2023)Li, Cheng, Zhao, Nie, and Wen}]{li2023halueval}
Junyi Li, Xiaoxue Cheng, Wayne~Xin Zhao, Jian-Yun Nie, and Ji-Rong Wen. 2023.
\newblock Halueval: A large-scale hallucination evaluation benchmark for large language models.
\newblock \emph{arXiv preprint arXiv:2305.11747}.

\bibitem[{Li et~al.(2021)Li, Ma, Yu, Wu, Gao, Ji, and McKeown}]{li-etal-2021-timeline}
Manling Li, Tengfei Ma, Mo~Yu, Lingfei Wu, Tian Gao, Heng Ji, and Kathleen McKeown. 2021.
\newblock \href {https://doi.org/10.18653/v1/2021.emnlp-main.519} {Timeline summarization based on event graph compression via time-aware optimal transport}.
\newblock In \emph{Proceedings of the 2021 Conference on Empirical Methods in Natural Language Processing}, pages 6443--6456, Online and Punta Cana, Dominican Republic. Association for Computational Linguistics.

\bibitem[{Lin(2004)}]{lin2004ROUGE}
Chin-Yew Lin. 2004.
\newblock Rouge: A package for automatic evaluation of summaries.
\newblock In \emph{Text summarization branches out}, pages 74--81.

\bibitem[{Martschat and Markert(2017)}]{martschat-markert-2017-improving}
Sebastian Martschat and Katja Markert. 2017.
\newblock \href {https://aclanthology.org/E17-2046} {Improving {ROUGE} for timeline summarization}.
\newblock In \emph{Proceedings of the 15th Conference of the {E}uropean Chapter of the Association for Computational Linguistics: Volume 2, Short Papers}, pages 285--290, Valencia, Spain. Association for Computational Linguistics.

\bibitem[{Min et~al.(2023)Min, Krishna, Lyu, Lewis, Yih, Koh, Iyyer, Zettlemoyer, and Hajishirzi}]{min-etal-2023-factscore}
Sewon Min, Kalpesh Krishna, Xinxi Lyu, Mike Lewis, Wen-tau Yih, Pang Koh, Mohit Iyyer, Luke Zettlemoyer, and Hannaneh Hajishirzi. 2023.
\newblock \href {https://doi.org/10.18653/v1/2023.emnlp-main.741} {{FA}ct{S}core: Fine-grained atomic evaluation of factual precision in long form text generation}.
\newblock In \emph{Proceedings of the 2023 Conference on Empirical Methods in Natural Language Processing}, pages 12076--12100, Singapore. Association for Computational Linguistics.

\bibitem[{Ng and Abrecht(2015)}]{ng-abrecht-2015-better}
Jun-Ping Ng and Viktoria Abrecht. 2015.
\newblock \href {https://doi.org/10.18653/v1/D15-1222} {Better summarization evaluation with word embeddings for {ROUGE}}.
\newblock In \emph{Proceedings of the 2015 Conference on Empirical Methods in Natural Language Processing}, pages 1925--1930, Lisbon, Portugal. Association for Computational Linguistics.

\bibitem[{Rajaby~Faghihi et~al.(2022)Rajaby~Faghihi, Alhafni, Zhang, Ran, Tetreault, and Jaimes}]{rajaby-faghihi-etal-2022-crisisltlsum}
Hossein Rajaby~Faghihi, Bashar Alhafni, Ke~Zhang, Shihao Ran, Joel Tetreault, and Alejandro Jaimes. 2022.
\newblock \href {https://doi.org/10.18653/v1/2022.findings-emnlp.400} {{C}risis{LTLS}um: A benchmark for local crisis event timeline extraction and summarization}.
\newblock In \emph{Findings of the Association for Computational Linguistics: EMNLP 2022}, pages 5455--5477, Abu Dhabi, United Arab Emirates. Association for Computational Linguistics.

\bibitem[{Setty(2024)}]{setty2024factcheck}
Vinay Setty. 2024.
\newblock Factcheck editor: Multilingual text editor with end-to-end fact-checking.
\newblock \emph{arXiv preprint arXiv:2404.19482}.

\bibitem[{Song et~al.(2024)Song, Chim, Tsakalidis, Ive, Atzil-Slonim, and Liakata}]{song-etal-2024-combining}
Jiayu Song, Jenny Chim, Adam Tsakalidis, Julia Ive, Dana Atzil-Slonim, and Maria Liakata. 2024.
\newblock \href {https://aclanthology.org/2024.findings-acl.873} {Combining hierachical {VAE}s with {LLM}s for clinically meaningful timeline summarisation in social media}.
\newblock In \emph{Findings of the Association for Computational Linguistics ACL 2024}, pages 14651--14672, Bangkok, Thailand and virtual meeting. Association for Computational Linguistics.

\bibitem[{Steen and Markert(2022)}]{steen-markert-2022-find}
Julius Steen and Katja Markert. 2022.
\newblock \href {https://aclanthology.org/2022.coling-1.527} {How to find strong summary coherence measures? a toolbox and a comparative study for summary coherence measure evaluation}.
\newblock In \emph{Proceedings of the 29th International Conference on Computational Linguistics}, pages 6035--6049, Gyeongju, Republic of Korea. International Committee on Computational Linguistics.

\bibitem[{Swan and Allan(2000)}]{swan2000automatic}
Russell Swan and James Allan. 2000.
\newblock Automatic generation of overview timelines.
\newblock In \emph{Proceedings of the 23rd annual international ACM SIGIR conference on Research and development in information retrieval}, pages 49--56.

\bibitem[{Tao et~al.(2024)Tao, Agrawal, Dombi, Sydorenko, and Lee}]{tao-etal-2024-chatgpt}
Yufei Tao, Ameeta Agrawal, Judit Dombi, Tetyana Sydorenko, and Jung~In Lee. 2024.
\newblock \href {https://aclanthology.org/2024.lrec-main.278} {{C}hat{GPT} role-play dataset: Analysis of user motives and model naturalness}.
\newblock In \emph{Proceedings of the 2024 Joint International Conference on Computational Linguistics, Language Resources and Evaluation (LREC-COLING 2024)}, pages 3133--3145, Torino, Italia. ELRA and ICCL.

\bibitem[{Team(2023)}]{team2023Qwen}
Qwen Team. 2023.
\newblock Qwen technical report.
\newblock \emph{arXiv preprint arXiv:2309.16609}.

\bibitem[{Tran et~al.(2013)Tran, Tran, Tran, Alrifai, and Kanhabua}]{tran2013leveraging}
Giang~Binh Tran, Tuan~A Tran, Nam-Khanh Tran, Mohammad Alrifai, and Nattiya Kanhabua. 2013.
\newblock Leveraging learning to rank in an optimization framework for timeline summarization.
\newblock In \emph{SIGIR 2013 Workshop on Time-aware Information Access (TAIA}.

\bibitem[{Wang et~al.(2022)Wang, Zhang, Zhang, Yang, Gao, Wu, Dong, He, Zhuo, Yang, Huang, Li, Wu, Lu, Zhu, Chen, Han, Pan, Wang, Wang, Wu, Zeng, Chen, Gan, and Zhang}]{fengshenbang}
Junjie Wang, Yuxiang Zhang, Lin Zhang, Ping Yang, Xinyu Gao, Ziwei Wu, Xiaoqun Dong, Junqing He, Jianheng Zhuo, Qi~Yang, Yongfeng Huang, Xiayu Li, Yanghan Wu, Junyu Lu, Xinyu Zhu, Weifeng Chen, Ting Han, Kunhao Pan, Rui Wang, Hao Wang, Xiaojun Wu, Zhongshen Zeng, Chongpei Chen, Ruyi Gan, and Jiaxing Zhang. 2022.
\newblock Fengshenbang 1.0: Being the foundation of chinese cognitive intelligence.
\newblock \emph{CoRR}, abs/2209.02970.

\bibitem[{Wang et~al.(2015)Wang, Cardie, and Marchetti}]{wang-etal-2015-socially}
Lu~Wang, Claire Cardie, and Galen Marchetti. 2015.
\newblock \href {https://doi.org/10.3115/v1/N15-1112} {Socially-informed timeline generation for complex events}.
\newblock In \emph{Proceedings of the 2015 Conference of the North {A}merican Chapter of the Association for Computational Linguistics: Human Language Technologies}, pages 1055--1065, Denver, Colorado. Association for Computational Linguistics.

\bibitem[{Wang et~al.(2016)Wang, Mehdad, Radev, and Stent}]{wang-etal-2016-low}
William~Yang Wang, Yashar Mehdad, Dragomir~R. Radev, and Amanda Stent. 2016.
\newblock \href {https://doi.org/10.18653/v1/N16-1008} {A low-rank approximation approach to learning joint embeddings of news stories and images for timeline summarization}.
\newblock In \emph{Proceedings of the 2016 Conference of the North {A}merican Chapter of the Association for Computational Linguistics: Human Language Technologies}, pages 58--68, San Diego, California. Association for Computational Linguistics.

\bibitem[{Wu and Hu(2018)}]{wu2018learning}
Yuxiang Wu and Baotian Hu. 2018.
\newblock Learning to extract coherent summary via deep reinforcement learning.
\newblock In \emph{Proceedings of the AAAI conference on artificial intelligence}, volume~32.

\bibitem[{Xu et~al.(2024)Xu, Fei, Pan, Liu, Lee, and Hsu}]{xu2024faithful}
Jundong Xu, Hao Fei, Liangming Pan, Qian Liu, Mong-Li Lee, and Wynne Hsu. 2024.
\newblock Faithful logical reasoning via symbolic chain-of-thought.
\newblock \emph{arXiv preprint arXiv:2405.18357}.

\bibitem[{Young et~al.(2024)Young, Chen, Li, Huang, Zhang, Zhang, Li, Zhu, Chen, Chang et~al.}]{young2024yi}
Alex Young, Bei Chen, Chao Li, Chengen Huang, Ge~Zhang, Guanwei Zhang, Heng Li, Jiangcheng Zhu, Jianqun Chen, Jing Chang, et~al. 2024.
\newblock Yi: Open foundation models by 01. ai.
\newblock \emph{arXiv preprint arXiv:2403.04652}.

\bibitem[{Zeng et~al.(2022)Zeng, Liu, Du, Wang, Lai, Ding, Yang, Xu, Zheng, Xia et~al.}]{zeng2022glm}
Aohan Zeng, Xiao Liu, Zhengxiao Du, Zihan Wang, Hanyu Lai, Ming Ding, Zhuoyi Yang, Yifan Xu, Wendi Zheng, Xiao Xia, et~al. 2022.
\newblock Glm-130b: An open bilingual pre-trained model.
\newblock \emph{arXiv preprint arXiv:2210.02414}.

\bibitem[{Zhang et~al.(2023)Zhang, Li, Cui, Cai, Liu, Fu, Huang, Zhao, Zhang, Chen et~al.}]{zhang2023siren}
Yue Zhang, Yafu Li, Leyang Cui, Deng Cai, Lemao Liu, Tingchen Fu, Xinting Huang, Enbo Zhao, Yu~Zhang, Yulong Chen, et~al. 2023.
\newblock Siren's song in the ai ocean: a survey on hallucination in large language models.
\newblock \emph{arXiv preprint arXiv:2309.01219}.

\bibitem[{Zhang et~al.(2024)Zhang, Cao, Ye, Ma, Liao, and Chua}]{zhang2024analyzingtemporalcomplexevents}
Zhihan Zhang, Yixin Cao, Chenchen Ye, Yunshan Ma, Lizi Liao, and Tat-Seng Chua. 2024.
\newblock \href {https://arxiv.org/abs/2406.02472} {Analyzing temporal complex events with large language models? a benchmark towards temporal, long context understanding}.
\newblock \emph{Preprint}, arXiv:2406.02472.

\end{thebibliography}

\appendix
\section{Event Atoms Decomposition}
Evaluation metrics based on event atoms require decomposition on both the reference and generated timelines. The event atoms in reference timelines are annotated in advance by human annotators. For the generated timelines, the decomposition process should be completed in real-time. Similar to the decomposition for atomic facts \citep{min-etal-2023-factscore}, we adopt GPT-3.5 to complete the automatic annotation.
\subsection{Manual Decomposition Protocol}
To maintain consistency and accuracy in event atom decomposition, human annotators should follow this protocol:

(1) Understanding the Context: Read the entire node summary carefully to understand the overarching event or narrative described and then identify the primary subject(s) and action(s) within the sentence.

(2) Segmentation of Events: Down each sentence into smaller units by identifying distinct actions or states that involve a subject and an object. Then, consider each clause within a complex sentence as a potential event atom if it represents a unique action or state. For instance, for the sentence ``John arrived at the station and met his friend.'', two event atoms can be identified:
\begin{itemize}
    \item Event Atom 1: ``John arrived at the station.''
    \item Event Atom 2: ``John met his friend.''
\end{itemize}

\subsection{Automatic Decomposition} \label{app:decompose}
The automatic decomposition process using GPT-3.5 is implemented by prompts in Table \ref{tab:atom}.

\begin{table*}[t]
    \centering

    \begin{tabularx}{1.0\textwidth}{|X|}
\toprule
\rowcolor{gray!25} 
\textbf{Atoms Decomposition Prompts}\\
\textbf{System Prompt}\\
\makecell[l]{
You are a Fact Decomposer. \\
\#\# Your task is: \\
As a specialized journalist, you will be provided with a sentence that may describe multiple events. \\Your task is to decompose the sentence into atomic propositions. An atomic proposition consists of, \\and only of, a subject, a predicate, and an object.\\
\#\# Output format: \\
Please use the following format for your output:\\
{[}``Atom\_{1}'', ``Atom\_{2}'', \dots{]}\\
\#\# Example:\\
Here is an example for you to better understand the task:\\
Input: ``Myanmar military: one-year state of emergency imposed''\\
Output: [``Myanmar military imposes state of emergency'', ``State of emergency lasts for one year'']}\\
\textbf{Input}\\
\{Node Summary of a timeline\}\\
\bottomrule
    \end{tabularx}    
    \caption{Prompts used for Atoms Decomposition Process. We show examples for the model to better comprehend the task. }
    \label{tab:atom}
\end{table*}

\section{Coherence Review Form} \label{app:coherence}
\begin{figure*}[ht]
    \centering
    \includegraphics[width=.95\textwidth]{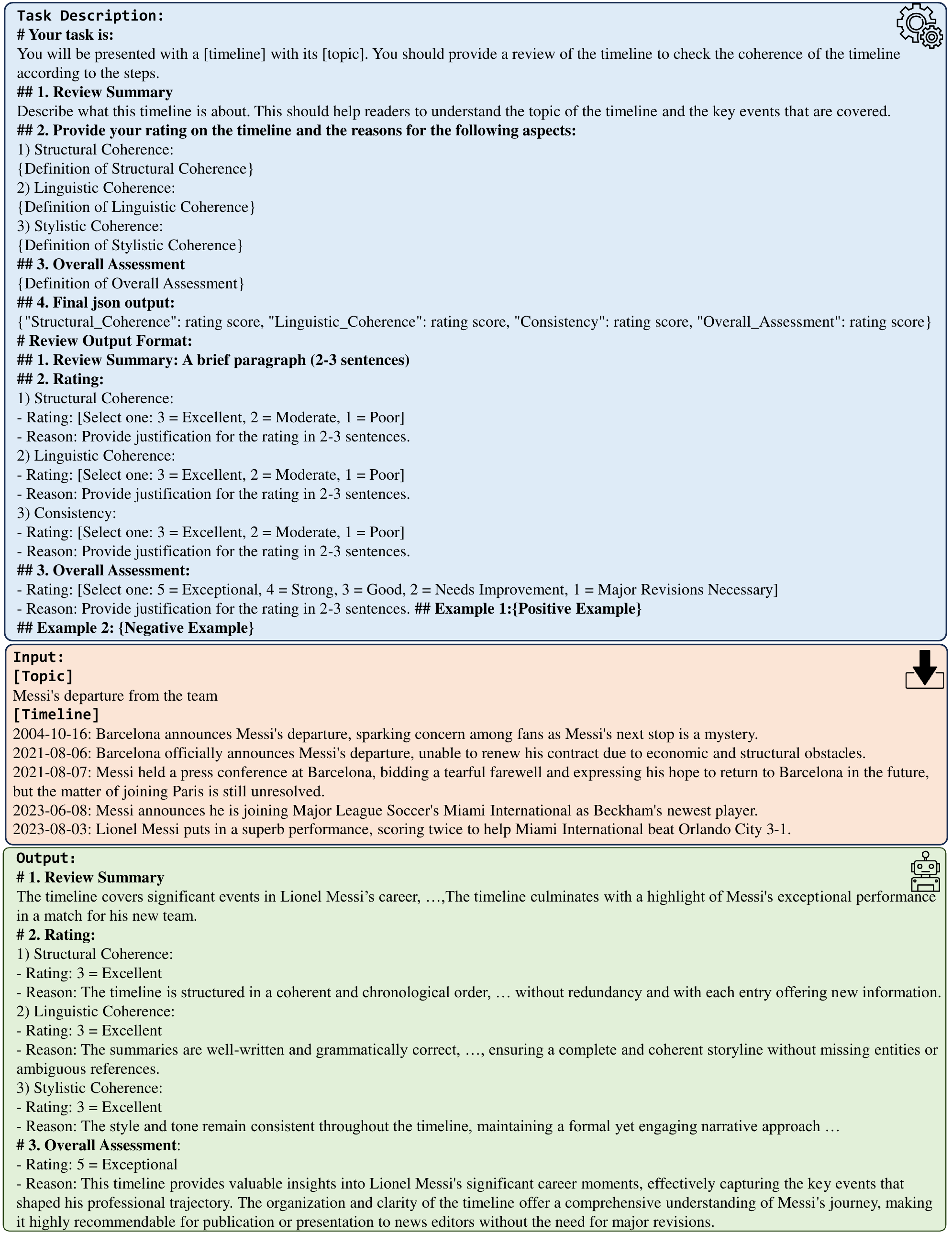}
    \caption{Example of the Coherence Review Form.}
    \label{fig:coherence}
\end{figure*}
\begin{figure*}[ht]
    \centering
    \includegraphics[width=.95\linewidth]{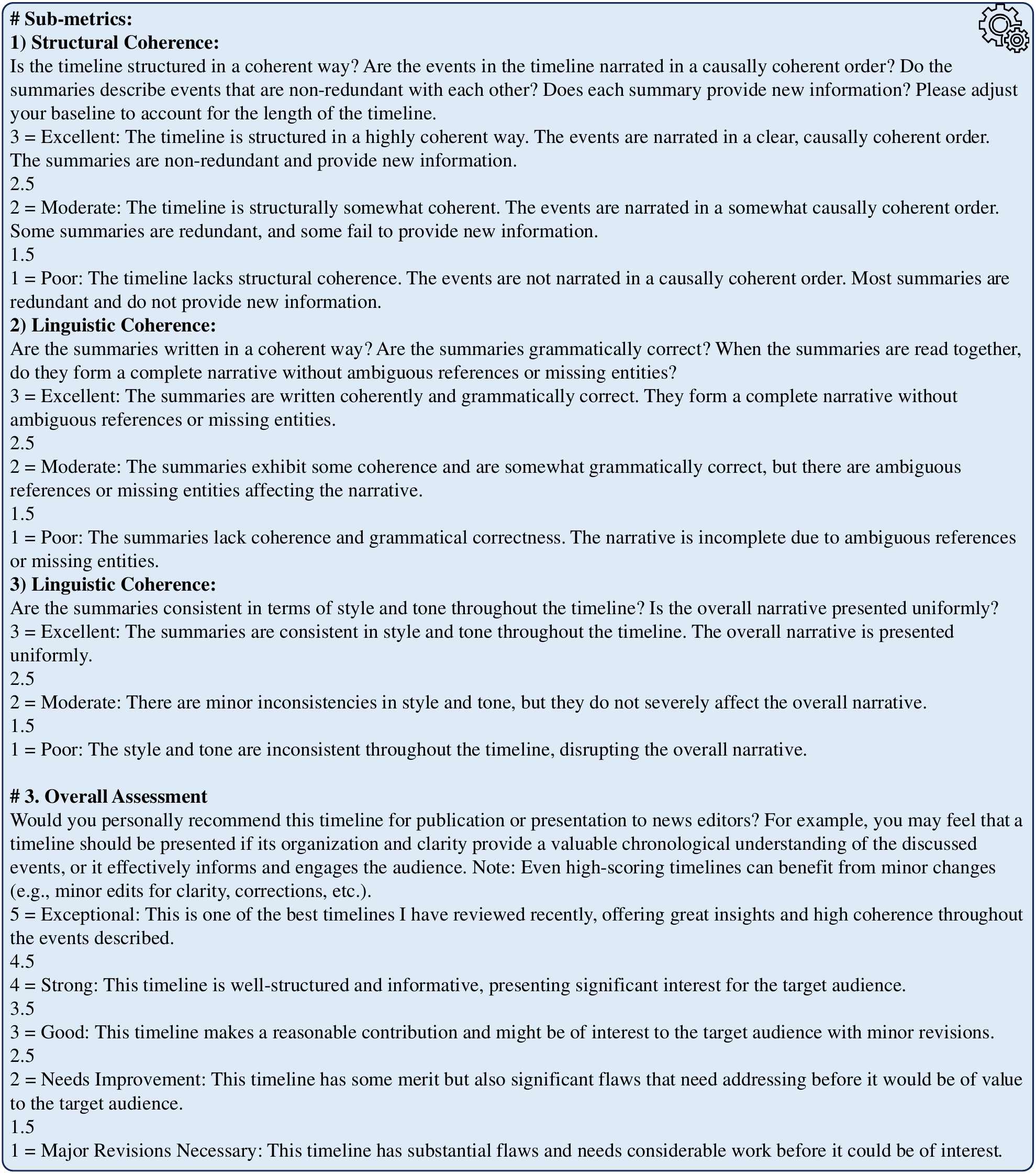}
    \caption{Sub-metrics and overall assessment definition with their corresponding score criteria.}
    \label{fig:coh_submatric}
\end{figure*}
In this section, we introduce the details of the Coherence Review Form and the sub-metrics.

Coherence is assessed through a review process similar to the ACL Review Form. As illustrated in Figure \ref{fig:coherence}, we decompose the review form into three steps. 
We ask experts to provide a comprehensive categorization of the main coherence errors that occur in timeline summarization. Based on these errors, we propose three sub-metrics along with their evaluation criteria for the scores (detailed in Figure \ref{fig:coh_submatric}).
We provide annotators with the task definition, detailed descriptions of the sub-metrics, and examples, both positive and negative, annotated by domain experts.
For automatic evaluation, we apply GPT-4o API and apply the \textit{Task Descriptions} as system prompts. 

\section{Details of Dataset Construction} \label{app:dataset}
In this section, we provide a detailed description of the dataset construction.
\subsection{Data Collection.}
 The data collection encompasses news topics of varying complexity, types, and scales. The original dataset contains 1,012 news topics. We apply a filtering standard for the news topics detailed as follows:  

\textbf{Diversity of Sources.}
To ensure a broad representation of perspectives and avoid bias in data collection,  we include timelines and articles sourced from a diverse set of reputable news sources. For the news topics: the dataset encompasses diverse news topics from various geographical regions across domestic and international events (a total of 32 countries/regions are included). For the timeline, the timeline nodes should come from multiple sources. Only timelines with events sourced from at least three news sources are included. For the news articles, we select articles from 2,858 sources including global press, forums, and social media, 

\textbf{Timeliness of News Topics.}The pretraining data for the models used in our evaluation are all prior to October 2023. To minimize the risk of using contaminated data from LLMs pretraining corpus, we exclude older or stale news topics that no longer reflect current events or have lost relevance over time. We select news topics after Oct. 2023 for dataset creation.

\textbf{Event-centric and Complete News Topics.}
To ensure that each news topic revolves around a specific, well-defined event or series of related events. We retain topics that provide comprehensive coverage of the development and conclusion of events, capturing key milestones and outcomes. We let annotators evaluate each selected topic to confirm it narrates a coherent storyline from beginning to end, avoiding fragmented or ambiguous narratives, and verify that the reference timelines associated with each topic adequately depict the chronology and significance of events.

By applying the standards, our dataset filter to 543 high-quality, multiple-sources news topics. Then, news articles for each topic are collected with the following steps:
(1) Retrieve on search engines (Baidu, Google, and Bing) for articles with news topics as keywords and time limits to the beginning and end of the corresponding timeline to get the most relevant 5 articles. (2) For each reference node summaries annotated in Baidu event websites, we directly get one source article from the website. Then we apply the summary as keywords on previously mentioned search engines to get 4 articles. (3) Filter low-quality articles. By thresholding the article titles against the news topics, we filter out low-quality articles. Articles with a BERT embedding similarity score of less than 0.3 between the title and the news topic are filtered out.

We list the statistics of the dataset in Figure \ref{fig:dataset}. 

\subsection{Consensus-based Annotation} \label{app:consensus}
To facilitate consensus, we prompt GPT-4o to play different roles as annotators, including news editor, journalist, and NLP researcher. These annotators focus on different aspects. The prompts are listed in Table \ref{tab:prompt}.

GPT-4o annotators are instructed with the decomposed  ``event atoms'' from the fine-grained timeline. These event atoms are grouped based on their timestamps, which could later be used to construct medium-grained and coarse-grained timelines.
To determine the consensus among GPT-4o agents, we employed the following approach:
(1) Each of the three GPT-4o agents independently selected their top 10 groups.
(2) Groups selected by all three agents are automatically included in the final selection.
(3) Groups selected by two agents are reviewed for inclusion based on their relevance and importance.
After the independent selections are made, the selected event atom groups from all three annotators are compared. The primary focus here is to identify the level of consensus among the annotators. We list the agreement degree of the annotated nodes as shown in Table \ref{tab:agreement}.
Once GPT-4 has selected the initial set of events, domain experts review and refine these selections to ensure accuracy and completeness. The refinement process includes:
(1) Fact-Checking: Ensuring that each selected event was factually accurate and well-supported by credible sources.
(2) Composing Atoms: Composing Atomic Facts into node summaries. 
(3) Coherence Refinement: Refine the summary in total to ensure that the timeline as a whole presents a coherent narrative.
(4) Detail Adjustment: Adding or removing details as necessary to meet the target granularity.

\begin{table*}
    \centering
    \small
    \begin{tabularx}{1.0\textwidth}{|X|}
        \toprule
        \rowcolor{gray!25} 

        \textbf{Role: News editor}\\ 
        \textbf{System Prompt}\\
        You are a specialized news editor. Your response should be in JSON format, start with ``\{'' and end with ``\}''.\\
        
        \textbf{Input}\\
        \makecell[l]{
        Given the event atom groups derived from the original timeline, your task is to select the most critical event groups that \\
        should be included in a condensed timeline. As a News Editor, focus on the following metrics:\\
        - Inclusive: The selected event groups should maximize the coverage of key events, ensuring that the most newsworthy and \\impactful events are included.\\
        - Accurate: The selected event groups must accurately reflect the essential developments without adding any ambiguity or \\misinformation.\\
        - Traceable: Ensure each selected event group can be directly traced back to the original timeline, maintaining the integrity \\and source of information.\\
        Please select the top \{N\} event atom groups according to the [\textbf{Input}]. Your response must follow the [\textbf{Template}].\\
        {[}\textbf{Example}{]}\\
        \{Example Annotation Results\}\\
        {[}\textbf{Template}{]}\\
        {[}``Group\textsubscript{1}'', ``Group\textsubscript{2}'', \dots, ``Group\textsubscript{\{N\}}''{]} \# Selected Event Atom Groups.\\
        {[}\textbf{Input}{]}\\
        Topic: \{Input Topic\}\\
        Event Atom Groups: \{Event Atom Groups of the timeline\}\\
        }\\
        \midrule
        \rowcolor{gray!25} 
        \textbf{Role: Journalist}\\ 
        \textbf{System Prompt}\\
        You are a specialized journalist. Your response should be in JSON format, start with ``\{'' and end with ``\}''.\\
        \textbf{Input}\\
        \makecell[l]{Given the event atom groups derived from the original timeline, your task is to select the most newsworthy event groups that \\should be included in a condensed timeline. As a News Editor, focus on the following metrics:\\
        - Insightfulness: Focus on selecting event atom groups that offer deep insights into the topic, providing the audience with a \\comprehensive understanding of the events' context and implications.\\
        - Objectivity: Ensure the selected groups are presented in an unbiased manner, maintaining journalistic integrity by avoiding \\sensationalism or subjective interpretation.\\
        - Relevance: Select event atom groups that are most relevant to the central theme or story, ensuring that the timeline remains \\focused and cohesive.\\
        Please select the top \{N\} event atom groups according to the [\textbf{Input}]. Your response must follow the [\textbf{Template}].\\
        {[}\textbf{Example}{]}\\
        \{Example Annotation Results\}\\
        {[}\textbf{Template}{]}\\
        {[}``Group\textsubscript{1}'', ``Group\textsubscript{2}'', \dots, ``Group\textsubscript{\{N\}}''{]} \# Selected Event Atom Groups.\\
        }\\
        \rowcolor{gray!25} 
        \midrule
        \textbf{Role: NLP Researcher}\\ 
        \textbf{System Prompt}\\
        You are a specialized NLP researcher. Your response should be in JSON format, start with ``\{'' and end with ``\}''.\\
        \textbf{Input}\\
        \makecell[l]{
        Given the event atom groups derived from the original timeline, your task is to select the most comprehensive event groups \\that should be included in a condensed timeline. As an NLP researcher, focus on the following metrics: \\
        - Comprehensiveness: Ensure that the selected event atom groups provide broad and detailed coverage of the original \\timeline, capturing all significant events and nuances.\\
        - Accuracy: Focus on selecting event atom groups that are factually correct, with a high level of precision in how the events \\are described, avoiding any distortion of the original data.\\
        - Reproducibility: Prioritize event atom groups that can be easily traced back to the original data, ensuring that the selections \\are well-documented and can be verified by others.\\
        Please select the top \{N\} event atom groups according to the [\textbf{Input}]. Your response must follow the [\textbf{Template}].\\
        {[}\textbf{Example}{]}\\
        \{Example Annotation Results\}\\
        {[}\textbf{Template}{]}\\
        {[}``Group\textsubscript{1}'', ``Group\textsubscript{2}'', \dots, ``Group\textsubscript{\{N\}}''{]} \# Selected Event Atom Groups.\\
        }\\
        \bottomrule

    \end{tabularx}
    \caption{Prompts for event consensus-based annotation, where N denotes the number of reference timeline nodes, in our case 10 and 5 for medium- and coarse-grained annotations.}
    \label{tab:prompt}

\end{table*}

\begin{table}[t]
\centering
\begin{tabular}{lcc}
\toprule
\textbf{Agreement Type} & \textbf{Count} & \textbf{Percentage} \\ 
\midrule
Full Agreement& 3118 & 45.09\% \\ 
\midrule 
Partial (1, 2)  & 2316 & 33.49\% \\ 
Partial (1, 3) & 573  & 8.28\%  \\ 
Partial (2, 3) & 380  & 5.50\%  \\ 
\midrule
No Agreement& 525  & 7.59\%  \\ 
\bottomrule
\end{tabular}
\caption{Agreement among annotators, where 1, 2, and 3 correspond to the news editor, journalist, and NLP researcher, respectively. The agreement can be categorized as: 
(1) Full Agreement: The annotators selected the same event atom group. (2) Partial Agreement: Two out of three annotators selected the same event atom group. (3) No Agreement: No common event atom groups were selected by the annotators.}
\label{tab:agreement}
\end{table}

\section{Methods Implementation} \label{app:method}
The extractive baselines are implemented based on the original code provided by the authors. For LLMs, we use the official API. For Qwen \citep{team2023Qwen}, we build upon their open-sourced weights\footnote{\url{https://huggingface.co/Qwen}}. The NLI model is implemented by BERT-based models \citep{devlin2018bert} fine-tuned on Chinese NLI datasets \citep{fengshenbang}. The evaluation metrics include informativeness (Info), granular consistency (Granu), factuality (Fact), and coherence.

\subsection{Details of LLM-based Methods}
We choose LLMs with advanced Chinese capability, including both open-source and closed-source models for our analysis. For closed-source LLMs, we select the most representative GPT series and widely known Chinese model Yi-medium. For open-source LLMs, we select GLM-3 and Qwen series with multiple model sizes. Particularly, 
for LP, we evaluate models including GPT-4o (128k) \citep{achiam2023gpt}, Yi-medium (200k) \citep{young2024yi} and GLM-3-Turbo (128k) \citep{zeng2022glm}. For hierarchical merging, we evaluate GPT-3.5-Turbo\footnote{\url{https://platform.openai.com/docs/models/GPT-3-5}}, GLM-3-Turbo \citep{zeng2022glm}, Yi-medium \citep{young2024yi}, and Qwen1.5 \citep{team2023Qwen}. The temperature is set to 0 for  greedy sampling.

\subsection{Long-congtext Prompting}
\begin{figure}[t]
    \centering
    \includegraphics[width=0.95\linewidth]{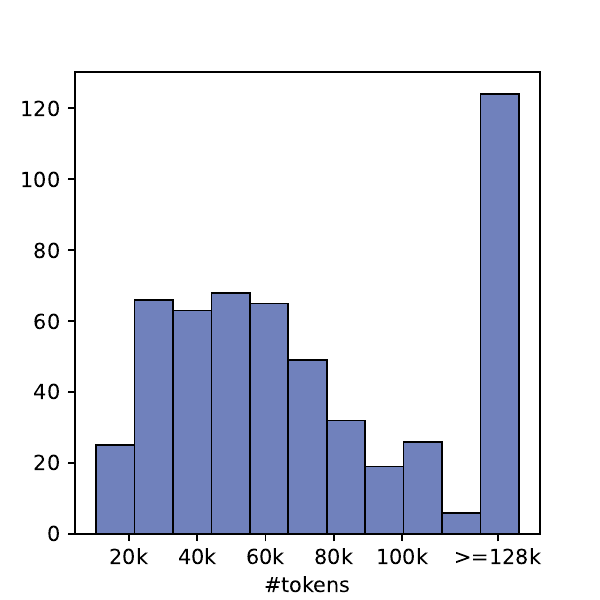}
    \caption{Token consumption histograms distribution for Long-context Prompting.}
    \label{fig:token}
\end{figure}

The prompts used for long-context prompting are illustrated in Table \ref{tab:long}. To handle topics with hundreds of articles that may exceed the maximum token length, we truncate the last paragraph of each article recursively until the total content falls within the token limit. Figure \ref{fig:token} shows the distribution of token consumption for long-context prompting.

\subsection{Hierarchical Merging}
The hierarchical merging method first generates a day summary for the news topic based on prompts in Table \ref{tab:hierarch}. Then, all nodes are hierarchically merged to form a complete timeline. Similarly, if the input exceeds the token length, we do the same operation as in long-context prompting.

\begin{table*}[t]
    \centering
    \small
    \begin{tabularx}{1.0\textwidth}{|X|}
\toprule
\rowcolor{gray!25} 
\textbf{Long-context Prompts}\\
\textbf{System Prompt}\\
\makecell[l]{
You are a News Event Timeline Generator. \\
\#\# Your task is: \\
As a specialized journalist, you will be provided with a news [topic] and related news [articles]. \\Based on this information, construct a chronologically ordered timeline summarizing the key events of the [topic]. \\Each event summary should be accompanied by an accurate timestamp. \\
\#\# Output format:\\ 
Please use the following format for your output:\\ 
1. yyyy-mm-dd: Event summary 1 \\
2. yyyy-mm-dd: Event summary 2 \\
\dots \\
\{N\}. yyyy-mm-dd: Event summary \{N\}\\ 
\#\# Note:\\ 
- The timeline should contain at least \{N\} event summaries.\\ 
- The summary content must match the timestamp.\\
- It's important to select key events to build the timeline, as not all [articles] are worth summarizing.
}\\
\textbf{Input}\\
\makecell[l]{
[Topic]\\
\{Topic of the Timeline\}\\
{[}Article 0{]}\\
Title: \{Title of Article 0\}\\
Release-time: \{Release-time of Article 0\}\\
Content: \{Content of Article 0\}\\
\dots\\
}\\
\bottomrule
    \end{tabularx}    
    \caption{Prompts used in Long-context Prompting (LP) for long-context large language models. Here, N denotes the required node amounts for the timeline.}
    \label{tab:long}
\end{table*}

\begin{table*}[t]
    \centering
\small
    \begin{tabularx}{1.0\textwidth}{|X|}
\toprule
\rowcolor{gray!25} 
\textbf{Day Summary Prompts}\\
\textbf{System Prompt}\\
\makecell[l]{
You are a News Event Timeline Generator.\\
\#\# Your task is:\\
As a specialized journalist, you will be provided with a news [topic] and related news [articles]. Based on this information, \\construct a chronologically ordered timeline summarizing the key events of the [topic]. Each event summary should be \\accompanied by an accurate timestamp.\\
\#\# Output format:\\ 
Please use the following format for your output:\\ 
1. yyyy-mm-dd: Event summary 1 \\
2. yyyy-mm-dd: Event summary 2 \\
\dots \\
\{N\}. yyyy-mm-dd: Event summary \{N\}\\ 
\#\# Note:\\ 
- There can only be ONE event summary per day.\\
- It's important to select key events to build the timeline, as not all [articles] are worth summarizing.
}\\
\textbf{Input}\\
\makecell[l]{
[Topic]\\
\{Topic of the Timeline\}\\
{[}Article 0{]}\\
Title: \{Title of Article 0\}\\
Release-time: \{Release-time of Article 0\}\\
Content: \{Content of Article 0\}\\
\dots
}\\
\midrule
\rowcolor{gray!25} 
\textbf{Timeline Merging Prompts}\\
\textbf{System Prompt}\\
\makecell[l]{
You are a News Event Timeline Generator. \\
\#\# Your task is:\\
As a specialized journalist, you will be provided with a news [topic], multiple partially completed timelines. Based on \\this information, merge the timelines to create a chronologically ordered timeline summarizing the key events of the [topic]. \\
\#\# Output format: \\
Please use the following format for your output: \\
1. yyyy-mm-dd: Event summary 1 \\
2. yyyy-mm-dd: Event summary 2 \\
\dots \\
N. yyyy-mm-dd: Event summary N\\ 
\#\# Note:\\ 
- There can only be ONE event summary per day.\\
- It's important to select key events to build the timeline, as not all events are worth summarizing.
}\\
\textbf{Input}\\
\makecell[l]{
[Timeline 0]\\
{Timeline 0}\\
\dots\\
}\\
\bottomrule
    \end{tabularx}    
    \caption{Prompts used in Hierarchical Merging (HM) for context length-limited large language models, where N denotes the required node amounts for the timeline.}
    \label{tab:hierarch}
\end{table*}

\subsection{Natural Language Granular Instruction}
We list the natural language granularity instruction in Table \ref{tab:natural_instruct}.

\begin{table*}
    \centering
    \setlength{\tabcolsep}{1mm} 
    
    \begin{tabularx}{1.0\linewidth}{l|c|X}
    \toprule
    Type  &\#Node & \multicolumn{1}{c}{$\mathcal{G}_o$} \\ 
    \midrule
    \multirow{2}{*}{Prompt} & 5 & \makecell[l]{Please generate a coarse-grained timeline.[Task Prompt*]}\\
    \cmidrule(lr){2-3}
     & N & \makecell[l]{Please generate a fine-grained timeline. [Task Prompt*]}\\
     \midrule
    \multirow{4}{*}{One-shot} & 5 & \makecell[l]{Please generate a timeline like:\\\{Timeline with 5 nodes\}}\\
    \cmidrule(lr){2-3}
     & N & \makecell[l]{Please generate a timeline like: \\\{Timeline with \{N\} nodes\}}\\
    \bottomrule
    \end{tabularx}
    \caption{Natural language granularity instructions used in the experiments. [Task Prompt*] denotes the prompts used for timeline summarization (LP and HM prompts in our paper). The ``*'' indicates that the node amounts \{N\} in the [Task Prompt] is replaced with ``N'' to represent an arbitrary number.}
    \label{tab:natural_instruct}
\end{table*}

\section{Influential Factors on Metrics} \label{app:exp}
\begin{figure}[t]
    \centering
    \includegraphics[width=1.0\linewidth]{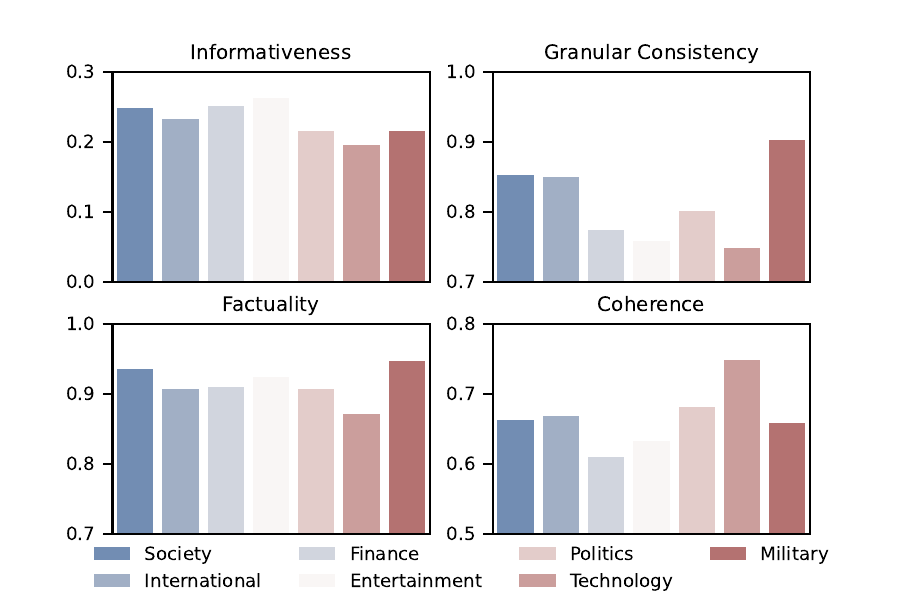}
    \caption{Topic types' influence on hierarchical merging GPT-3.5-Turbo.} 
    \label{fig:topic-type}
\end{figure}

We analyze the influence of topic types on the performance of hierarchical merging with GPT-3.5-Turbo. The results are shown in Figure \ref{fig:topic-type}. 
We observe that the performance on different topic types varies significantly. ``Military'' topics consistently achieve the highest scores in all metrics, suggesting that the model handles structured and well-defined content more effectively. Conversely, ``Politics'' and ``Technology'' topics present the greatest challenges, particularly in informativeness and coherence, likely due to the complexity and variability of information required in these domains.
This suggests that the model's performance is closely tied to the nature of the topic.

\begin{figure}[t]
    \centering
    \includegraphics[width=1.0\linewidth]{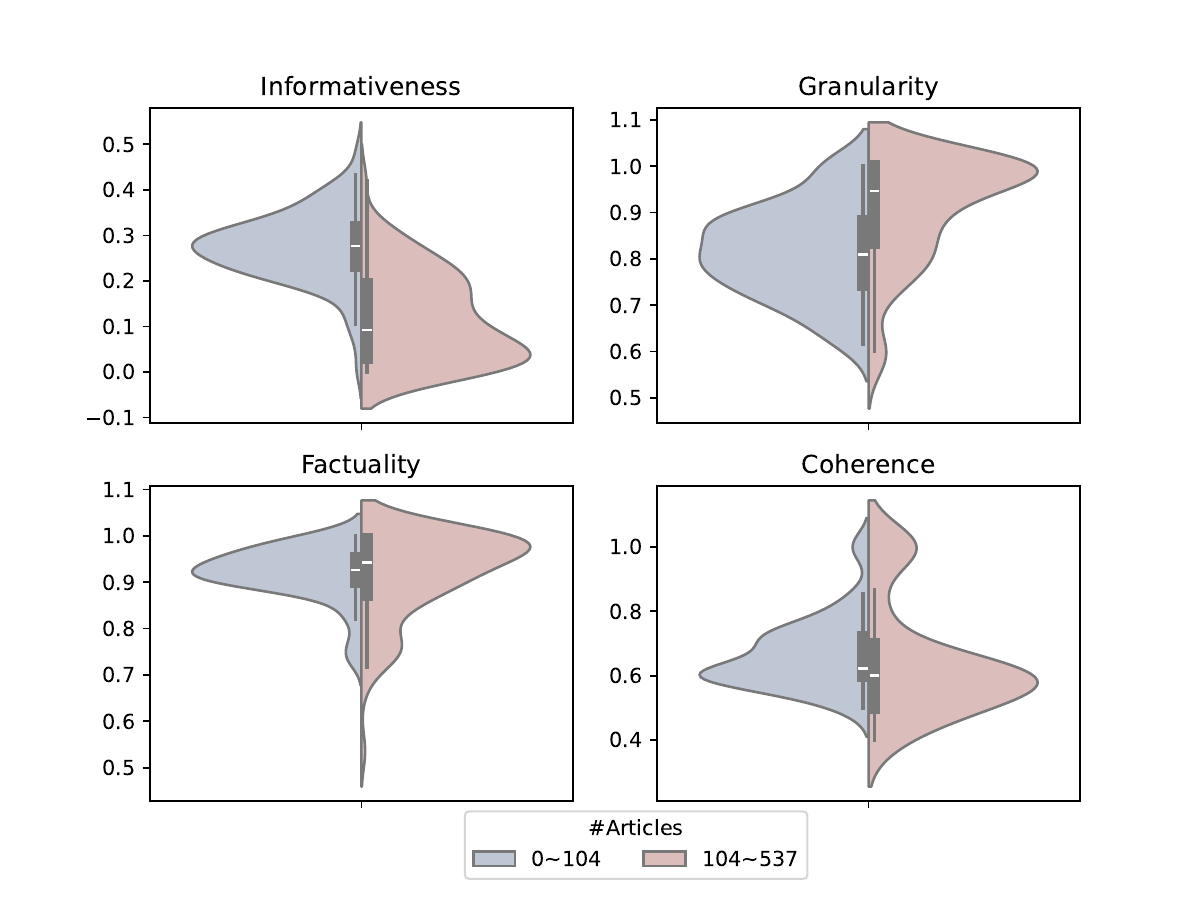}
    \caption{The influence of the number of news articles on evaluation metrics.}
    \label{fig:news_num}
\end{figure}

We also assess the influence of the number of news articles. The results are shown in Figure \ref{fig:news_num}. We find that the model performs better with fewer news articles, as the model can better capture the key information and generate more coherent summaries. However, the factuality of the summaries decreases with fewer news articles, as the model may lack sufficient information to generate faithful summaries.

\section{Details of Alignment Evaluation} \label{app:alignment}
To measure how well human evaluators' assessments of timelines align with the proposed metrics.
\subsection{Evaluation Process}
\begin{figure*}
    \centering
    \includegraphics[width=0.95\linewidth]{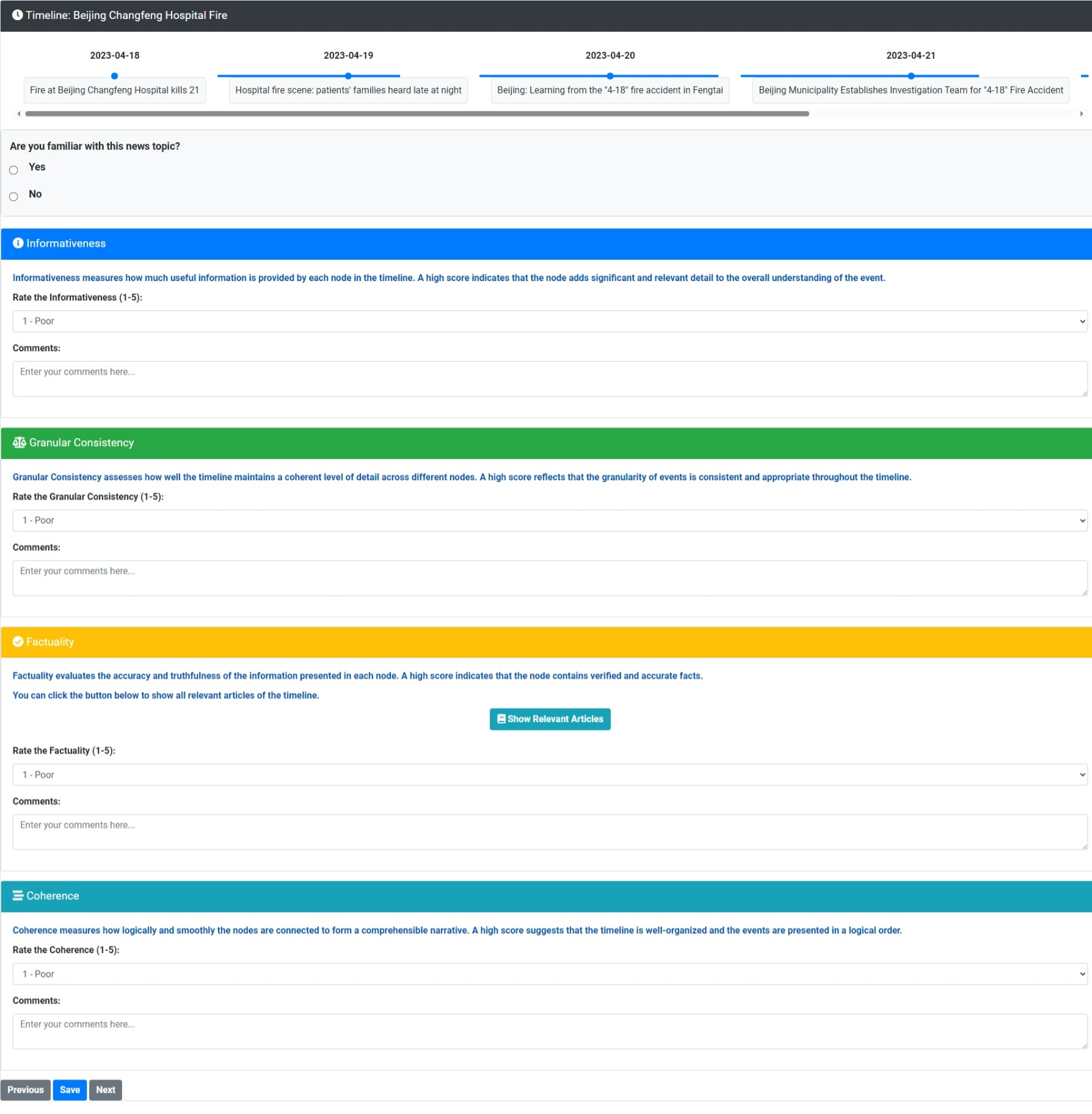}
    \caption{Human annotation scoring sheets of the proposed metrics.}
    \label{fig:human_1}
\end{figure*}

We choose three evaluators with a background in journalism and experience in summarization or timeline construction. Then, we prepare a set of 50 timelines generated by your DTELS system. Include a mix of high and low scores across different dimensions. We have each evaluator independently assess the timelines using the scoring sheets in Figure \ref{fig:human_1}. After the initial round, we facilitate a group discussion where evaluators can compare their scores and discuss discrepancies. This can help in understanding different perspectives and potentially refining the evaluation criteria. Then, we allow evaluators to revise their scores based on insights gained from the discussion.

\subsection{Metrics Definitions}

\subsubsection{Informativeness}
Informativeness measures how much useful information is provided by each node in the timeline. A high score indicates that the node adds significant and relevant detail to the overall understanding of the event.

\subsubsection{Granular Consistency}
Granular Consistency assesses how well the timeline maintains a coherent level of detail across different nodes. A high score reflects that the granularity of events is consistent and appropriate throughout the timeline.

\subsubsection{Factuality}
Factuality evaluates the accuracy and truthfulness of the information presented in each node. A high score indicates that the node contains verified and accurate facts.

\subsubsection{Coherence}
Coherence measures how logically and smoothly the nodes are connected to form a comprehensible narrative. A high score suggests that the timeline is well-organized and the events are presented in a logical order.

\end{document}